\documentclass[runningheads]{llncs}
\usepackage{graphicx}
\usepackage{amsmath,amssymb} %
\usepackage{color}\usepackage[width=122mm,left=12mm,paperwidth=146mm,height=193mm,top=12mm,paperheight=217mm]{geometry}

\usepackage{hyperref}

\usepackage{xspace}
\usepackage{multirow}
\usepackage{algorithmic,algorithm}
\usepackage{eucal}

\def\sh{{StructHash}\xspace}
\def\shh{{StructHash}\xspace}

\newcommand{\comment}[1]{}

\newcommand{\st}{{\rm s.t.\!:}\xspace}

\newcommand{\ones}{{\mathbf{1}}}

\def\T{{\!\top}}

\def\w{{\boldsymbol w}}

\def\blambda{{\boldsymbol \lambda}}

\def\mslack{$ m $-slack }
\def\1slack{$1$-slack }

\def\v{{\boldsymbol v}}
\def\c{{\boldsymbol c}}
\def\x{{\boldsymbol x}}
\def\w{{\boldsymbol w}}
\def\y{{\boldsymbol y}}

\def\xb{{\boldsymbol x}}
\def\yb{{\boldsymbol y}}
\def\xib{{\boldsymbol \xi}}

\def\hx{  \widetilde{\boldsymbol x} }

\DeclareMathOperator{\Scal}{\mathcal{S}}

\DeclareMathOperator{\Wcal}{\mathcal{W}}
\DeclareMathOperator{\Xcal}{\mathcal{X}}
\DeclareMathOperator{\Ycal}{\mathcal{Y}}

\def\Real{\mathbb{R}}
\def\argmax{\operatorname*{argmax\,}}

\def\loss{{\it \Delta}}
\def\dpsi{{\delta \psi}}

\def\argmax{\operatorname*{argmax\,}}

\let\Transpose\T

\newcommand{\figcenter}[1]{\raisebox{-0.5\height}{#1}}
\newcommand{\colseperator}{\hspace*{.1in}}

\newcommand{\eq}[1]{(\ref{#1})}

\begin{document}

\pagestyle{headings}
\mainmatter
\title{Optimizing Ranking Measures for Compact Binary Code Learning\thanks{Appearing in Proc.\ European Conf. Computer
Vision, 2014.}} %

\titlerunning{Optimizing Ranking Measures for Compact Binary Code Learning}

\authorrunning{Lin et al.}

\author{Guosheng Lin$^1$, Chunhua Shen$^1$\thanks{Corresponding author (e-mail: chunhua.shen@adelaide.edu.au).},
  Jianxin Wu$^2$
 }
\institute{$^1$University of Adelaide, Australia ~ ~ ~$^2$Nanjing University, China}

\maketitle

\begin{abstract}

Hashing has proven  a valuable tool for large-scale information retrieval. Despite
much success, existing hashing methods optimize over simple objectives such as
the reconstruction error or graph Laplacian related loss functions, instead of the
performance evaluation criteria of interest---multivariate performance measures
such as the AUC and NDCG. Here we present a general framework (termed
StructHash) that allows one to directly optimize multivariate performance
measures. The resulting optimization problem can involve exponentially or infinitely many
variables and constraints, which is more challenging than standard structured output
learning. To solve the StructHash optimization problem, we use a combination of
column generation and cutting-plane techniques. We demonstrate the generality
of StructHash by applying it to ranking prediction and image retrieval, and
show that it outperforms a few state-of-the-art hashing methods.

\end{abstract}

\section{Introduction}
\label{sec:Intro}

    The ever increasing volumes of imagery available, and the benefits
    reaped through the interrogation of large image datasets, have
    increased enthusiasm for large-scale approaches to vision.  One of
    the simplest, and most effective means of improving the scale and
    efficiency of an application has been to use hashing to
    pre-process the data
\cite{kulis2009learning,weiss2008spectral,ICCV13Lin,CVPR13aShen,liu2011hashingGraphs,CVPR14Lin}.

    Depending on applications, specific  measures are used to
    evaluate the performance of the generated hash codes. For example,
    information retrieval and ranking criteria \cite{mcfee10_mlr} such as
    the Area Under the ROC Curve (AUC) \cite{Joachims2005},
    Normalized Discounted Cumulative Gain (NDCG)
    \cite{JarvelinK00},
    Precision-at-K, Precision-Recall and Mean Average Precision (mAP) have been widely
    adopted to evaluate the success of hashing methods. However, to date,
    most hashing methods are usually learned by optimizing simple
    errors such as the reconstruction error (e.g., binary
    reconstruction embedding hashing  \cite{kulis2009learning}) or
    the graph Laplacian related loss
    \cite{zhangSTHs,liu2011hashingGraphs,weiss2008spectral}.
    These hashing methods construct a set of hash functions that map the
    original high-dimensional data into a much smaller binary space,
    typically with the goal of preserving neighborhood relations.
     The resulting compact binary encoding enables fast similarity
     computation between data points by using the Hamming distance,
     which can be carried out by rapid, often hardware-supported,
     bit-wise operations.  Furthermore, compact binary codes are much
     more efficient for large-scale data storage.

    To our knowledge, none of the existing hashing methods have
    tried to learn hash codes that {\em directly} optimizes a multivariate
    performance criterion.
    In this work, we seek to reduce the discrepancy between existing learning criteria and
the evaluation criteria (such as retrieval quality measures).

    The proposed framework accommodates
    various complex multivariate measures.
    By observing that the hash codes learning problem is essentially
    an information retrieval problem, various ranking loss functions
    can and should be applied, rather than merely pairwise distance comparisons.
    This framework also allows to introduce more general definitions of
    ``similarity'' to hashing beyond existing ones.

    In summary, our main contributions are as follows.
    \begin{enumerate}
    \item
    We propose a flexible binary hash codes learning framework that
    directly optimizes complex multivariate measures. This framework, {\em for
    the first time}, exploits the gains made in structured output learning for
    the purposes of hashing.  Our hashing method, labelled as \shh,
    is able to directly
    optimize various multivariate evaluation criteria, such as information
    retrieval measures (e.g., AUC and
     NDCG \cite{JarvelinK00}).

    \item
    To facilitate \shh, we combine column generation and cutting-plane methods
    in order to efficiently solve the resulting optimization problem,
    which may involve exponentially or even infinitely many variables
    and constraints.

    \item
    Applied to ranking prediction for image retrieval, the proposed method demonstrates
    state-of-the-art performance on hash function learning.

    \end{enumerate}

    \section{Related work}
    One of
    the best known {\em data-independent}  hashing methods is locality sensitive
    hashing (LSH)
    \cite{Gionis1999}, which uses random projection to generate binary
    codes.
    Recently,  a number of {\em data-dependent}
    hashing methods have been proposed. For example,  spectral hashing (SPH)
    \cite{weiss2008spectral} aims to preserve the neighbourhood
    relation by optimizing the Laplacian affinity.
    Anchor graph hashing (AGH)
    \cite{liu2011hashingGraphs} makes the original SPH much more scalable.
    Examples of supervised or semi-supervised hashing methods include
    binary reconstruction embedding (BRE)
    \cite{kulis2009learning}, which aims to minimise the expected distances;
    and the semi-supervised sequential
    projection learning hashing (SPLH) \cite{wang2010semi},
    which enforces the smoothness of similar data
    points and the separability of dissimilar data points.

To obtain a richer representation,  kernelized LSH \cite{KLSH} was proposed, which
randomly samples training data as support vectors, and randomly draws the dual
coefficients from a Gaussian distribution. Liu et al.\ extended Kulis and Grauman's work to
kernelized supervised hashing (KSH) \cite{KSH} by learning the dual coefficients instead.
Lin et al.\ \cite{CVPR14Lin} employed ensembles of decision trees as the hash functions.
Nonetheless, all of these methods  do not directly optimize the multivariate performance measures of interest.
    We formulate hash codes learning as a structured output
    learning problem, in order to directly optimize a wide variety of evaluation measures.

    We are primarily inspired by recent advances in learning to rank such as the metric learning method in
    \cite{mcfee10_mlr}, which directly optimizes several different
    ranking measures.  We aim to learn hash
    functions, which leads to a very different learning task preventing directly applying techniques in \cite{mcfee10_mlr}.
    We are also inspired by the recent column generation
    based hashing method, column generation hashing (CGH) \cite{ICML13a},
    which iteratively learns hash functions using
    column generation. However, their method optimizes
    the conventional classification-related loss, which is much
    simpler than the multivariate loss that we are interested in here.
    Moreover, the optimization of CGH relies on all triplet constraints
    while our method is able to use much less number of constraints without sacrificing the performance.

    Our framework is built on the structured SVM \cite{Tsochantaridis2004}, which has been
applied to many applications for complex structured output prediction, e.g., image
segmentation, action recognition and so on.

    \textbf{Notation}
    Let $\{( \xb_i; \yb_i ) \}$, $ i = 1, 2 \cdots$, denote a set of input and output pairs.
    The discriminative function for structured output prediction is\footnote{To be precise,
    the prediction function should be written as
    $F(\xb,\yb; \w)$ because it is parameterized by $ \w$. For simplicity, we omit $ \w $.
    }
    $F(\xb,\yb):
    \Xcal\times\Ycal\mapsto \Real$, which measures the compatibility of the input and output pair
    $(\xb,\yb)$.
    Given a query $ \xb_i $, we use $ \Xcal_i^+  $ and $ \Xcal_i^-  $
    to denote the subsets of relevant and irrelevant data points in
    the training data.
    Given two data points: $\x_i$ and $\x_j$,
     $\x_i \!\!   \prec_{\y} \!\!  \x_j $
     ($\x_i \!\! \succ_{\y} \!\! \x_j $) means that $\x_i $ is placed before
     (after) $\x_j $ in the ranking $\y$.

     \subsection{Structured SVM}
	First we provide a brief overview on structured SVM.
     Structured SVM enforces that the score of the
     ``correct'' model $ \yb$ should be larger than all other ``incorrect''
     model $ \yb' $,  $ \forall \yb' \neq \yb $, which writes:
     \begin{equation}
         \label{EQ:svm1}
         \forall \yb' \in \Ycal:\quad  %
            \w^\T [ \psi (  \xb, \yb  )  -
                    \psi (  \xb, \yb'  ) ]
                    \geq \loss( \yb, \yb'  ) - \xi.
     \end{equation}
     Here $  \xi $ is a slack variable (soft margin) corresponding to the hinge
     loss.
     $ \psi( \xb, \yb   )  $ is a vector-valued joint feature mapping.
     It plays a key role in structured learning and specifies the
     relationship between an input $ \xb $ and output $ \yb $. $\w$ is the model parameter. The label loss $\loss(\yb,\yb') \in \Real$ measures the discrepancy of the predicted $\yb'$ and the true
    label $\yb$.
    A typical assumption is that $\loss(\yb, \yb)=0, \loss(\yb, \yb')>0$ for any $\yb'\neq\yb$,
    and $\loss(\yb, \yb')$ is upper bounded.
     The prediction $\yb^*$ of an input $\x$ is achieved by
      \begin{align}
      \label{eq:infer1}
       \yb^*=\argmax_{\yb}F(\xb, \yb) = \w ^\T \psi( \xb, \yb ).
     \end{align}

    \subsubsection{Optimization for the model parameter $ \w $}

    For structured problems, the size of the output $|\Ycal|$ is typically very large.
    Considering all possible constraints in \eqref{EQ:svm1} is generally intractable.
    The cutting-plane method \cite{kelley1960} is commonly employed, which
    allows to maintain a small working-set of constraints and obtain an approximate solution of the original problem up to a pre-set precision.  To speed up, a 1-slack reformulation is proposed
    \cite{JoachimsSVM}.
    Nonetheless the cutting-plane method needs to find the most violated
    label (equivalent to an inference problem)
      \begin{align}
        \label{EQ:SSVM2}
        \argmax_{ \yb' \in \Ycal   }  \; \w^\Transpose \psi ( \xb, \yb'  )
        + \loss( \yb, \yb'  ).
    \end{align}
    Structured SVM typically requires: 1) a well-designed feature representation $ \psi(\cdot, \cdot)
    $; 2) an appropriate label loss $ \loss( \cdot, \cdot  ) $;
    3) solving inference problems \eq{eq:infer1} and \eq{EQ:SSVM2} efficiently.
    \subsubsection{Ranking prediction with structured output}

    In a retrieval system, given a test data point $\x$, the goal is to predict a ranking of data points in the database. For a
    ``correct'' ranking, relevant data points are expected to be
    placed in front of irrelevant data points. A ranking output is
    denoted by $\y$.  Let us introduce a symbol  $ y_{jk} = 1  $ if
    $\x_j \!\!   \prec_{\y} \!\!  \x_k $ and $ y_{jk} = -1 $
    if  $\x_j \!\!   \succ_{\y} \!\!  \x_k $.
    The ranking can be evaluated by various measures e.g. AUC, NDCG,
    mAP. These evaluation measures can be
    optimized directly as label losss
    $\loss$ \cite{Joachims2005,mcfee10_mlr}.
    Here $\psi(\x, \y)$ can be defined as:
  \begin{align}
	\psi(\xb_i, \yb)
	= %
	\sum_{\xb_j \in \Xcal_i^+} \sum_{\xb_k \in \Xcal_i^-}
    y_{jk}
    \Bigl[
        \frac{\phi( \xb_i, \xb_j  )  -  \phi( \xb_i, \xb_k )
        }
        {
             |\Xcal_i^+| \cdot | \Xcal_i^- |
        }
    \Bigr].
    \label{EQ:rk1}
  \end{align}
$\Xcal_i^+$ and $\Xcal_i^-$ are the sets of relevant and irrelevant
neighbours of data point $\x_i$ respectively.
Here $  | \cdot | $ is the set size.
The feature map $ \phi( \xb_i, \xb_j  )   $ captures the relation between a query $ \xb_i $ and point $ \xb_j $.

        We have briefly reviewed how to
        optimize ranking criteria using structured prediction. Now we review some
        basic concepts of hashing before introducing our framework.

\subsection{Learning-based hashing} %
\label{SUBSEC:Hashing}

        Given a  set of training data $ \xb_i $, ($ i = 1, 2, \dots$),
        the task is to learn a set of hash functions
        $ [ h_1( \xb  ), h_2 ( \xb), \dots, h_\ell(\xb)  ]  $. Each hash
        function maps the input into a binary bit $ \{0, 1 \} $.
        So
        with the learned functions, an input $ \xb $ is mapped into a
        binary code of length $ \ell $.
        We use $ \hx \in
        \{0, 1\}^\ell$ to denote the
        hashed values of $  \xb  $ i.e.,
        \begin{equation}
         \hx  =  [ h_1( \xb  ) \dots, h_\ell(\xb)     ]^\Transpose.
        \label{EQ:Hash}
        \end{equation}
        Suppose that we are given the supervision information as a set
        of triplets: $\{(\x_i, \x_j, \x_k)\} (i=1,2,\dots) $, in which
        $\x_j$ is an relevant (similar) data point of
        $\x_i$ (i.e.,  $ \xb_j \in \Xcal_i^+$) and
        $\x_k$ is an irrelevant
        (dissimilar) neighbor of $\x_i$ ( i.e., $ \xb_k \in
        \Xcal_i^-$).
        These triplets encode the relative similarity information.
        After applying the hashing, the distance of two hash
        codes (of length $\ell$) can be calculated using the weighted hamming distance:
        \begin{equation}
            \label{EQ:Hw}
             d_{\sf hm}(\x_i, \x_j )= \w^\T| \hx_i -\hx_j |,
        \end{equation}
        where $\w > 0 $ is a non-negative vector, which can be learned.
        Such weighted hamming distance is used in CGH \cite{ICML13a} and
        multi-dimension spectral hashing \cite{MDSH}.
        It is expected that after hashing, the distance between
        relevant data points should be smaller than the distance between
        irrelevant data points. That is
\[
    d_{\sf hm} (\x_i, \x_j) \leq  d_{\sf hm}(\x_i, \x_k),
\]
   for $  \x_j\in\Xcal_i^+, \x_k\in\Xcal_i^-, \forall i=1,2,\cdots. $
   One can then define the margin $ \rho = d_{\sf hm}(\x_i, \x_k)  -
   d_{\sf hm} (\x_i, \x_j)  $. It is then  possible to plug  this
   margin into the large-margin learning framework to optimize for the
   parameter $ \w $ as well as the hash functions, as shown in \cite{ICML13a}.

\section{The proposed \shh algorithm}

    For the time being, let us assume that we have already learned all
    the hashing functions. In other words, {\em given a data point $ \xb $,
    we assume that we have access to its corresponding hashed values $ \hx  $,
    as defined in \eqref{EQ:Hash}.
    }
    Later we will show how this mapping can be explicitly learned
    using column generation. Now let us focus on how to optimize for
    the weight $ \w $.
    When the weighted hamming distance is used, we aim to learn an
    optimal weight $ \w $ defined in \eqref{EQ:Hw}.
    Distances are calculated in the learned space and ranked
    accordingly.
    A
    natural choice for the vector-valued mapping function $ \phi $ in Equ.\
    \eqref{EQ:rk1} is
    \begin{equation}
        \label{EQ:phi}
         \phi( \xb_i, \xb_j   )  =  -  |  \hx_i - \hx_j   |.
    \end{equation}
    Note that we have changed the sign, which preserves the ordering
    in the standard structured SVM.
    Due to this change of sign, sorting the data by ascending $
    d_{\sf hm} ( \x_i, \x_j   )    $ is equivalent to sorting
    by descending $ \w^\Transpose  \phi ( \xb_i, \xb_j   ) = -
    \w^\Transpose | \hx_i - \hx_j | $.

    The loss function $ \loss( \cdot, \cdot ) $ depends on the
    metric, which we will discuss in detail in the next section.
    For ease of exposition, let us define
    \begin{equation}
        \label{EQ:delta}
        \delta  \psi_i ( \yb )
              = \psi ( \x_i, \yb_i ) - \psi ( \x_i, \yb ).
    \end{equation}
    We consider the following problem,
   \begin{subequations}
       \label{EQ:SHH3}
    \begin{align}
          \min_{ {\w \geq 0, \xib \geq 0}}   \;\; &
    \|\w \|_1 +  {\tfrac{C}{m}} \, \sum_{i=1}^m  \xi_{i} \\
    \st  \;\;
	&
    \forall i=1, \dots, m
    \text{ and } \forall \yb \in \Ycal: \notag \\
    &
    \w^\T \delta \psi_i ( \yb )  \geq
         \loss(\yb_i,\yb) - \xi_i. \label{eq:sh_con}
	\end{align}
	\label{eq:sh}
\end{subequations}
    Unlike standard structured SVM, here we use the
    $ \ell_1 $ regularisation (instead of $ \ell_2 $) and enforce that $ \w $ to be non-negative.
    This is aligned with boosting methods
    \cite{demiriz2002linear,Shen2011Totally}, and enables us to learn hash functions efficiently.

\subsection{Learning weights $\w$ via cutting-plane} %
\label{SUBSEC:OptW}
Here we show how to learn $\w$.
Inspired by \cite{JoachimsSVM},
we first derive the \1slack formulation of the original \mslack  formulation \eqref{eq:sh}:
\begin{subequations}
  \label{eq:sh-1s}
\begin{align}
  & \min_{\w \geq 0, \xi \geq 0}  \;\;
  \| \w \|_1 + C \xi \\
  \st  \;\;
  &
  \forall \c \in\{0,1\}^m \text{ and }
  \forall \y \in \Ycal, i=1,\cdots, m: \notag \\
  &
   \frac{1}{m} \w^ \T
  \biggl[
             \sum_{i=1}^m c_i
            \cdot
                    \dpsi_i ( \y )
    \biggr]
            \geq
            \frac{1}{ m }  \sum_{i=1}^m c_i\loss(\y_i,\y ) - \xi.
  \label{eq:sh-1s_con}
\end{align}
\end{subequations}
Here $ \c $ enumerates all possible $ \c \in \{ 0, 1 \}^m$.
As in \cite{JoachimsSVM}, cutting-plane methods can
    be used to solve the \1slack primal problem
    \eqref{eq:sh-1s} efficiently.
    Specifically, we need to solve a maximization for every $\x_i$
    in each cutting-plane iteration to find the most violated
    constraint of \eqref{eq:sh-1s_con},
given a solution $\w$:
\begin{align}
\y_i^*= \argmax_{\y} \loss( \y_i, \y ) - \w ^ \T  \dpsi_i (\y ).
\label{eq:infer}
\end{align}
We now know how to efficiently learn $\w$
using cutting-plane methods. However, it remains unclear how to learn
hash functions (or features). Thus far, we have taken for granted that
the hashed values $ \hx $ (or $h$) are given.
    We would like to learn the hash functions and $ \w $
    in a single optimization framework.  Next we show how this is
    possible using the column generation technique from boosting.

\begin{algorithm}[t]
\caption{\footnotesize \shh: Column generation for hash function learning}
\footnotesize{
1: {\bf Input:} training examples $ (\x_1; \y_1), (\x_2; \y_2) ,\cdots $;
parameter $C$; the maximum iteration number (bit length $ \ell$).

    2: {\bf Initialise:} working set of hashing functions $\Wcal_{\mathrm{H}}\leftarrow \emptyset$;
    initialise $\lambda_{(\c,\y)} = C/m $ by randomly picking $m$ pairs of $(\c,\y)$ and the rest is set to 0.

    3: {\bf Repeat}

    4:$\quad-$ Find a new hashing function $h^* ( \cdot )$ by solving
    Equ.\ \eqref{EQ:CG-Sub}.

	5:$\quad-$ add $h^*$ into the working set of hashing functions: $\Wcal_{\mathrm{H}}$.

    6:$\quad-$
    Solve the structured SVM problem \eqref{EQ:SHH3} or the equivalent
    \eqref{eq:sh-1s}
    using cutting-plane as discussed in Sec. \ref{SUBSEC:OptW}.

    7: {\bf Until}
        the maximum iteration is reached.

    8: {\bf Output:}
    Learned hash functions and $ \w $.
    }
\label{ALG:alg1}
\end{algorithm}

\subsection{Learning hash functions using column generation} %
\label{SUBSEC:CG}

 Note that the dimension of $ \w $ is the same as
 the dimension of $ \hx $ (and of $  \phi ( \cdot, \cdot )$, see Equ.\
 \eqref{EQ:phi}), which is the number of hash bits by the definition
 \eqref{EQ:Hash}. If we were able to
 access all hash functions, it may be possible to select a subset of
 them and learn the corresponding $ \w $ due to the sparsity introduced by the $ \ell_1 $
 regularization in \eqref{EQ:SHH3}.
    Unfortunately,
    the number of possible hash functions can be infinitely
    large. In this case it is in general infeasible to solve the optimization
    problem exactly.

    Column generation \cite{demiriz2002linear} can be used to approximately solve the problem
    by adding variables iteratively into the master optimization
    problems.
    Column generation was originally invented to solve extremely
    large-scale linear programming problem, which
    mainly works on the dual problem. The
basic concept of  column generation  is to add one constraint at a
time to the dual problem until an optimal solution is identified.
Columns\footnote{A column is a variable in the primal and a
corresponding constraint in the dual.}
are generated and added to the
problem iteratively to approach the optimality. In the primal problem, column
generation solves the problem on a subset of primal variables ($ \w $
in our case), which
corresponds to a subset of constraints in the dual.
This strategy has been widely employed to learn weak learners in boosting
\cite{Shen2011Totally,demiriz2002linear}.

 To learn hash functions via column generation, we derive the dual problem of
 the above \1slack optimization, which is,
\begin{subequations}
      \label{eq:sh-1s-dual}
  \begin{align}
    \max_{ \blambda \geq 0 }
    \;\;
    &
    \sum_{ \c , \y}
    \lambda_{ (\c, \y ) }
    \sum_{i=1}^m c_i  \loss( \y_i, \y ) \\
   \st \;\;
   &  \frac{1}{m} \sum_{ \c, \y } \lambda_{ (\c, \y) }
     \biggl[
          \sum_{i=1}^m c_i \cdot  \dpsi_i( \y )
     \biggr]
     \leq \ones, \label{EQ:11b}
\\
      &  0 \leq \sum_{ \c, \y} \lambda_{ (\c, \y ) } \leq C.
\end{align}
\end{subequations}
We denote by $ \lambda_{( \c, \y)} $ the \1slack dual variable associated with one constraint in \eqref{eq:sh-1s_con}.
Note that \eqref{EQ:11b} is a set of constraints
    because $ \delta \psi( \cdot )$ is a vector of the same dimension
    as $ \phi(\cdot, \cdot)  $ as well as $  \hx  $, which can be infinitely large.
One dimension in the vector $ \delta \psi( \cdot )$ corresponds to one constraint in \eqref{EQ:11b}.
Finding the most violated constraint in the dual form
\eqref{eq:sh-1s-dual} of the \1slack formulation for
    generating one hash function is to maximise the l.h.s.\ of
    \eqref{EQ:11b}.

    The calculation of $ \delta \psi( \cdot )$ in \eqref{EQ:delta} can be simplified as follows.
    Because of the subtraction of $\psi( \cdot )$ (defined in \eqref{EQ:rk1}),
    only those incorrect ranking pairs will appear in the calculation.
    Recall that the true ranking is $ \yb_i $ for $ \xb_i $.
    We define $ \Scal_i(\y ) $ as a set of incorrectly ranked pairs:
    $ (j, k) \in \Scal_i(\y )   $, in which the incorrectly ranked pair $(j, k)$ means
    that the true ranking is $  \xb_j \!\! \prec_{\y_i} \!\!  \xb_k  $
    but $ \xb_j \! \succ_{\y}  \! \xb_k  $. So we have
    \begin{align}
            \delta \psi_i ( \yb  )
               & =
               \tfrac{2}{ | \Xcal^+_i | |\Xcal^-_i | }
               \sum_ {
               (j,k) \in \Scal_i(\y)
               }
\bigl[
            \phi( \xb_i, \xb_j ) - \phi( \xb_i, \xb_k )
\bigr] \notag
    \\
    &           =
               \tfrac{2}{ | \Xcal^+_i | |\Xcal^-_i | }
               \sum_ {
               (j,k) \in \Scal_i(\y)
               }
\bigl(
| \hx_i - \hx_k | - |  \hx_i - \hx_j |
\bigr).
\end{align}
With the above equations and the definition of $\hx$ in \eqref{EQ:Hash},
the most violated constraint in  \eqref{EQ:11b}
can be found by solving the following problem:
\begin{align}
    \label{EQ:CG-Sub-tmp}
    h^* ( \cdot )  =  & \argmax_ { h(\cdot) }
    \sum_{\c, \yb } \lambda_{ (\c, \y ) }
    \sum_i
      \frac{2 c_i}{ | \Xcal^+_i | |\Xcal^-_i | }
      \cdot
   \notag    \\
     & \sum_ { (j,k) \in \Scal_i(\y) }
\bigl(
| h( \x_i ) - h(\x_k) | - |  h(\x_i) - h(\x_j) |
\bigr).
\end{align}
By exchanging the order of summations,
the above optimization can be further written in a compact form:
\begin{align}
    \label{EQ:CG-Sub}
    h^* ( \cdot ) &  =  \argmax_ { h(\cdot) } \sum_{i, \y}
     \sum_ { (j,k) \in \Scal_i(\y)  } \mu_{(i, \y)} \,
\bigl( | h( \x_i ) - h(\x_k) | - |  h(\x_i) - h(\x_j) |\bigr), \\
 & \text{where, }
	\mu_{(i, \y)} = \frac{2}{ | \Xcal^+_i | |\Xcal^-_i | } \sum_{\c} \lambda_{ (\c, \y ) } c_i.
\end{align}
The objective in the above optimization is a summation of weighted triplet $(i,j,k)$ ranking scores,
in which $\mu_{(i, \y)}$ is the triplet weighting value.
Solving the above optimization provides the best hash function
            for the current solution $\w$.
            Once a hash function is generated, we learn $ \w $ using cutting-plane in Sec. \ref{SUBSEC:OptW}.
The column generation procedure for hash function learning is
summarised in Algorithm \ref{ALG:alg1}.

The form of hash function $h(\cdot)$ can be any function that have binary output value. For a decision stump as the hash function,
usually we can exhaustively enumerate all possibility and find the globally best one. However globally solving \eqref{EQ:CG-Sub} is generally difficult. In our experiments, we use the linear perceptron hash function with the output in $\{0, 1\}$:
\begin{align}
\label{eq:hash_learn}
h(\x)=0.5( {\rm sign}(\v^\T\x+ b)+1).
\end{align}
The non-smooth function $ \rm sign(\cdot) $  here brings the difficulty for optimization.
Similar to \cite{ICML13a}, we replace the $ \rm sign(\cdot) $ function by a smooth
sigmoid function, and then locally solve the above optimization \eqref{EQ:CG-Sub} (e.g.,
LBFGS \cite{lbfgs}) for learning the parameters of a hash function.
We can apply a few heuristics to initialize for solving \eqref{EQ:CG-Sub}. For example,
similar to LSH, we can generate a set of random projection planes and then choose the best
one that maximizes the objective in \eqref{EQ:CG-Sub} as the initialization. We
can also train a decision stump by searching a best dimension and threshold to maximize the
objective on the quantized data. Alternatively, one can employ the spectral relaxation
method \cite{liu2011hashingGraphs} which drops the $ \rm sign(\cdot) $ function and solves
a generalized eigenvalue problem to obtain an initial point.
In our experiments, we use the spectral relaxation method for initialization.

Next, we discuss some widely-used information retrieval evaluation
criteria, and show how they can be seamlessly incorporated into \shh.

\section{Ranking measures} %
\label{SEC:RM}

Here we discuss
a few ranking measures for loss functions, including AUC, NDCG, Precision-at-K, and mAP.
Following \cite{mcfee10_mlr}, we define the loss function over two rankings $\loss \in [0 \; 1]$ as:
\begin{align}
\loss(\y,\y')=1-\text{score}(\y,\y').
\end{align}
Here $\y'$ is the ground truth ranking and $\y$ is the prediction. We define $\Xcal_{\y'}^+$ and $\Xcal_{\y'}^-$ as the indexes of relevant and irrelevant neighbours respectively in the ground truth ranking $\y'$.

{\bf AUC}. The area under the ROC curve
is to evaluate the performance of correct ordering of data pairs,
which can be computed by counting the proportion of correctly ordered data pairs:
\begin{align}
\mathrm{score}_{\mathrm{AUC}}(\y,\y')=\frac{1}{|\Xcal_{\y'}^+||\Xcal_{\y'}^-|}\sum_{i\in\Xcal_{\y'}^+}\sum_{j\in\Xcal_{\y'}^-}\delta(i \prec_\y j).
\end{align}
$\delta(\cdot) \in \{0, 1\}$ is the indicator function.
For using this AUC loss,
the maximization inference in \eqref{eq:infer} can be solved efficiently by sorting the distances of data pairs, as described in \cite{Joachims2005}.
Note that the loss of a wrongly ordered pair is not related to their positions in the ranking list, thus AUC is a position insensitive measure.

{\bf Precision-at-K}.
Precision-at-K is to evaluate the quality of top-K retrieved examples in a ranking. It is computed by counting the number of relevant data points within top-K positions and divided by K:
\begin{align}
\text{score}_{\mathrm{P@K}}(\y,\y')=\frac{1}{K}\sum_{i=1}^{K}\delta(\y(i) \in \Xcal^+).
\end{align}
Here $\y(i)$ is the example index on the $i$-th position of a ranking $\y$;
$\delta(\cdot)$ is an indicator.
An algorithm for solving the inference in
\eqref{eq:infer} is proposed in \cite{Joachims2005}.

{\bf NDCG}. Normalized Discounted Cumulative Gain \cite{JarvelinK00} is to measure the ranking quality of the first K returned neighbours. A similar measure is Precision-at-K which is the proportion of top-K relevant neighbours. NDCG is a position-sensitive measure which considers the positions of the top-K relevant neighbours.
Compared to the position-insensitive measure: AUC, NDCG assigns different importances on the ranking positions,
which is a more favorable measure for a general notion  of a ``good'' ranking in real-world applications.
In NDCG, each position of the ranking is assigned a score in a decreasing way. NDCG can be computed by accumulating the scores of top-K relevant neighbours:
\begin{align}
\mathrm{score}_{\mathrm{NDCG}}(\y, \y')=\frac{1}{\sum_{i=1}^{K}{S(i)}}\sum_{i=1}^{K}{S(i)}\delta(\y(i) \in
\Xcal_{\y'}^+).
\end{align}
Here $\y(i)$ is the example index on the $i$-th position of a ranking $\y$.
$S(i)$ is the score assigned to the $i$-th position of a ranking.
$S(1)=1$, $S(i)=0$ for $i>k$ and $S(i)=1/\log_2(i)$ for other cases.
A dynamic programming algorithm is proposed in \cite{Chakrabarti2008} for solving the maximization inference in \eqref{eq:infer}.

{\bf mAP}. Mean average precision (mAP) is the averaged precision-at-K scores over all positions of relevant data points in a ranking, which is computed as:
\begin{align}
\mathrm{score}_{\mathrm{mAP}}(\y, \y')=\frac{1}{|\Xcal_{\y'}^+|}\sum_{i=1}^{|\Xcal_{\y'}^+|+ |\Xcal_{\y'}^-|}\delta(\y(i)\in \Xcal_{\y'}^+) \mathrm{score}_{\mathrm{P@K}(K=i)}(\y, \y').
\end{align}
For using this mAP loss, an efficient algorithm for solving the inference in
\eqref{eq:infer} is proposed in \cite{Yue2007}.

\begin{table*}[t]
\caption{Results using NDCG measure (64 bits).
 We compare our StructHash using AUC (StructH-A) and NDCG (StructH-N) loss functions with other supervised and un-supervised methods.
 Our method using NDCG loss performs the best in most cases.}
\centering
\resizebox{.9\linewidth}{!}
  {
  \begin{tabular}{ c || c c | c c c c | c c c c c}
\hline
Dataset & StructH-N & StructH-A &CGH  &SPLH &STHs &BREs &ITQ  &SPHER  &MDSH &AGH  &LSH \\
\hline
& \multicolumn{11}{c}{NDCG ($K=100$)} \\ \hline
STL10 &\bf 0.435  &0.374  &0.375  &0.404  &0.214  &0.289  &0.337  &0.318  &0.313  &0.310  &0.228\\
USPS  &\bf  0.905 &0.893  &0.900  &0.816  &0.688  &0.777  &0.804  &0.762  &0.735  &0.741  &0.668\\
MNIST &0.851  &0.798  &\bf 0.867  &0.804  &0.594  &0.805  &0.856  &0.806  &0.100  &0.793  &0.561\\
CIFAR &0.335  &0.259  &0.258  &\bf 0.357  &0.178  &0.273  &0.314  &0.297  &0.283  &0.286  &0.168\\
ISOLET  &\bf 0.881  &0.839  &0.866  &0.629  &0.766  &0.483  &0.623  &0.518  &0.538  &0.536  &0.404\\
\hline
  \end{tabular}
  }
\label{tab:main}
\end{table*}

\begin{figure*}[t]
    \centering
   \includegraphics[width=.32\linewidth]{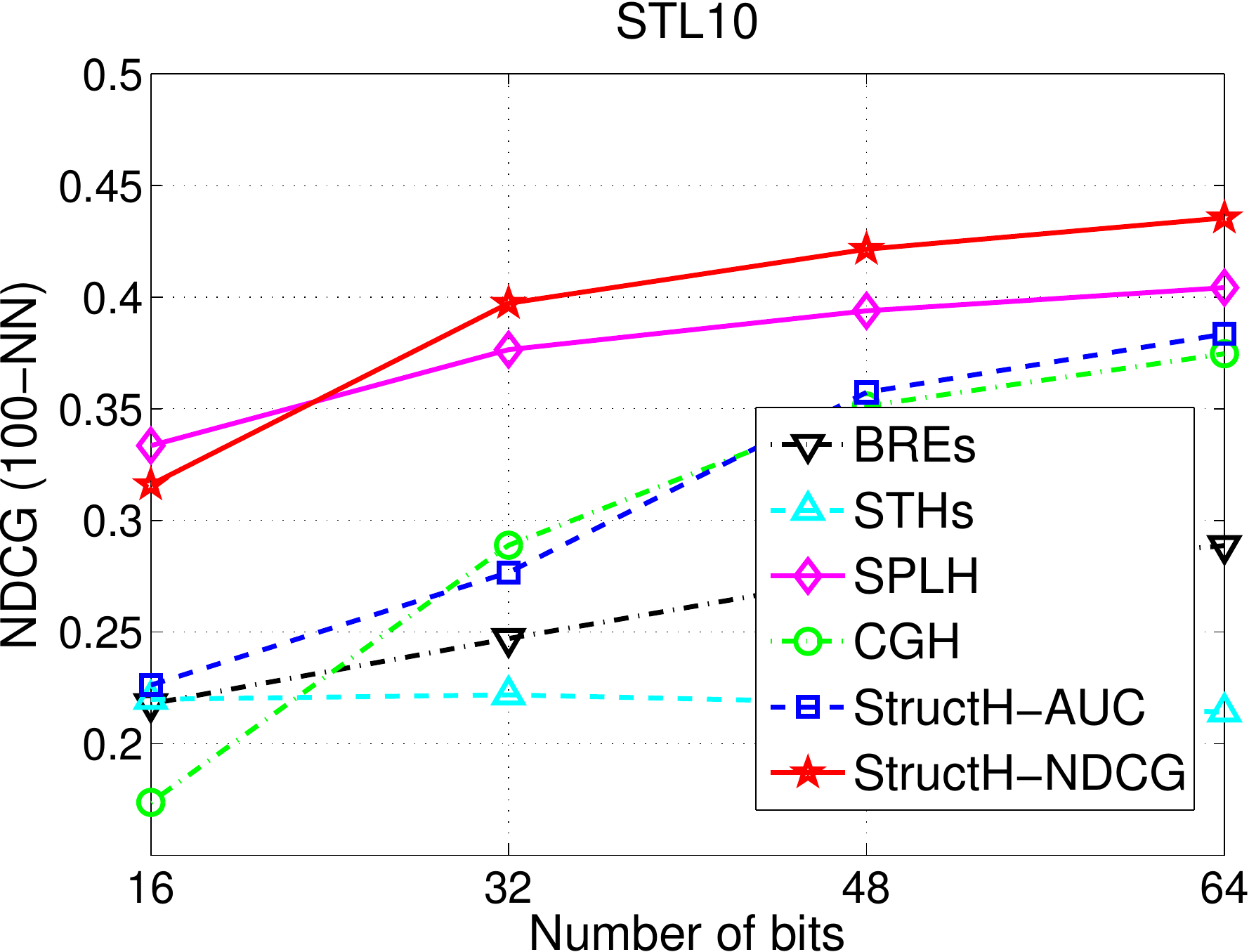}
   \includegraphics[width=.32\linewidth]{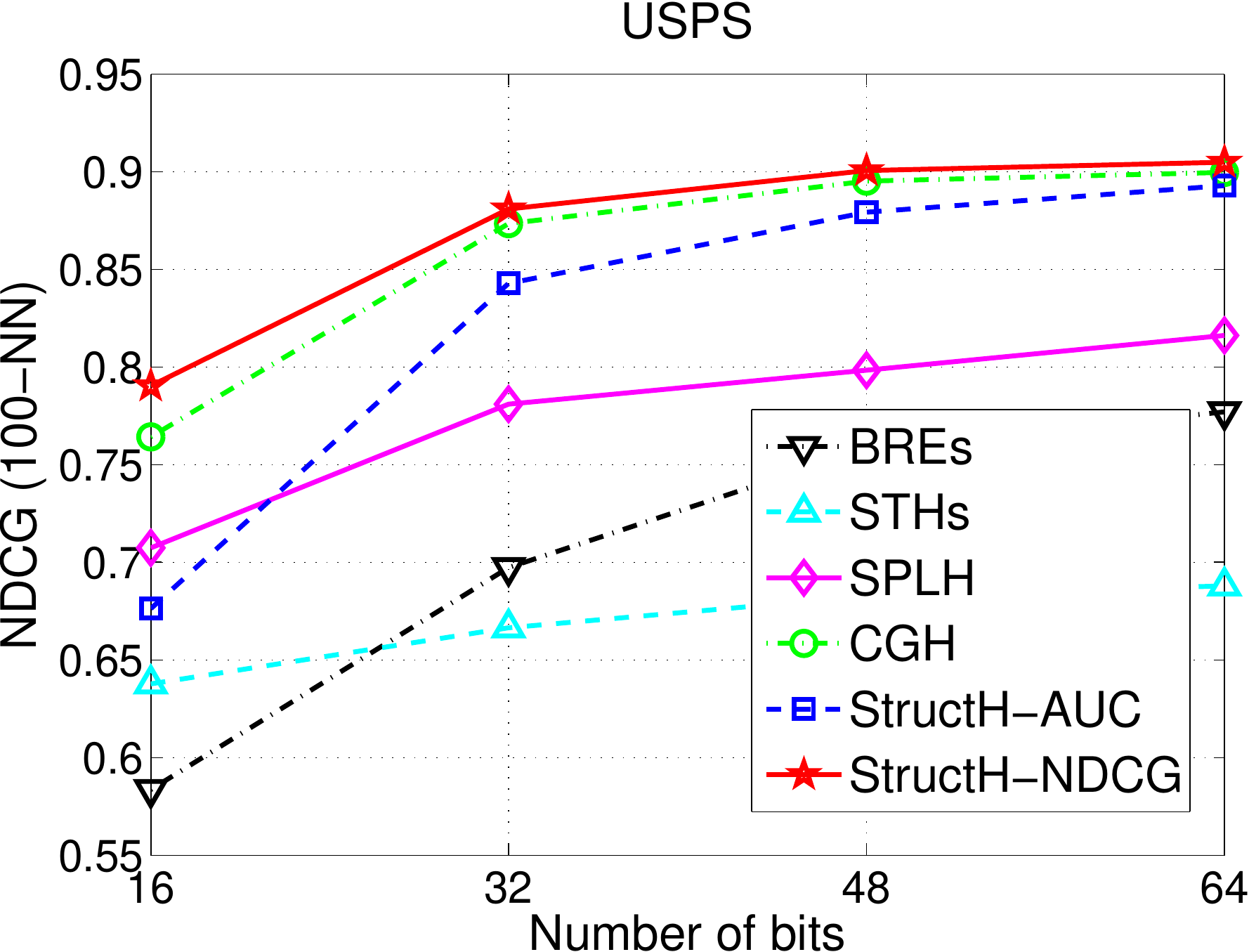}
   \includegraphics[width=.32\linewidth]{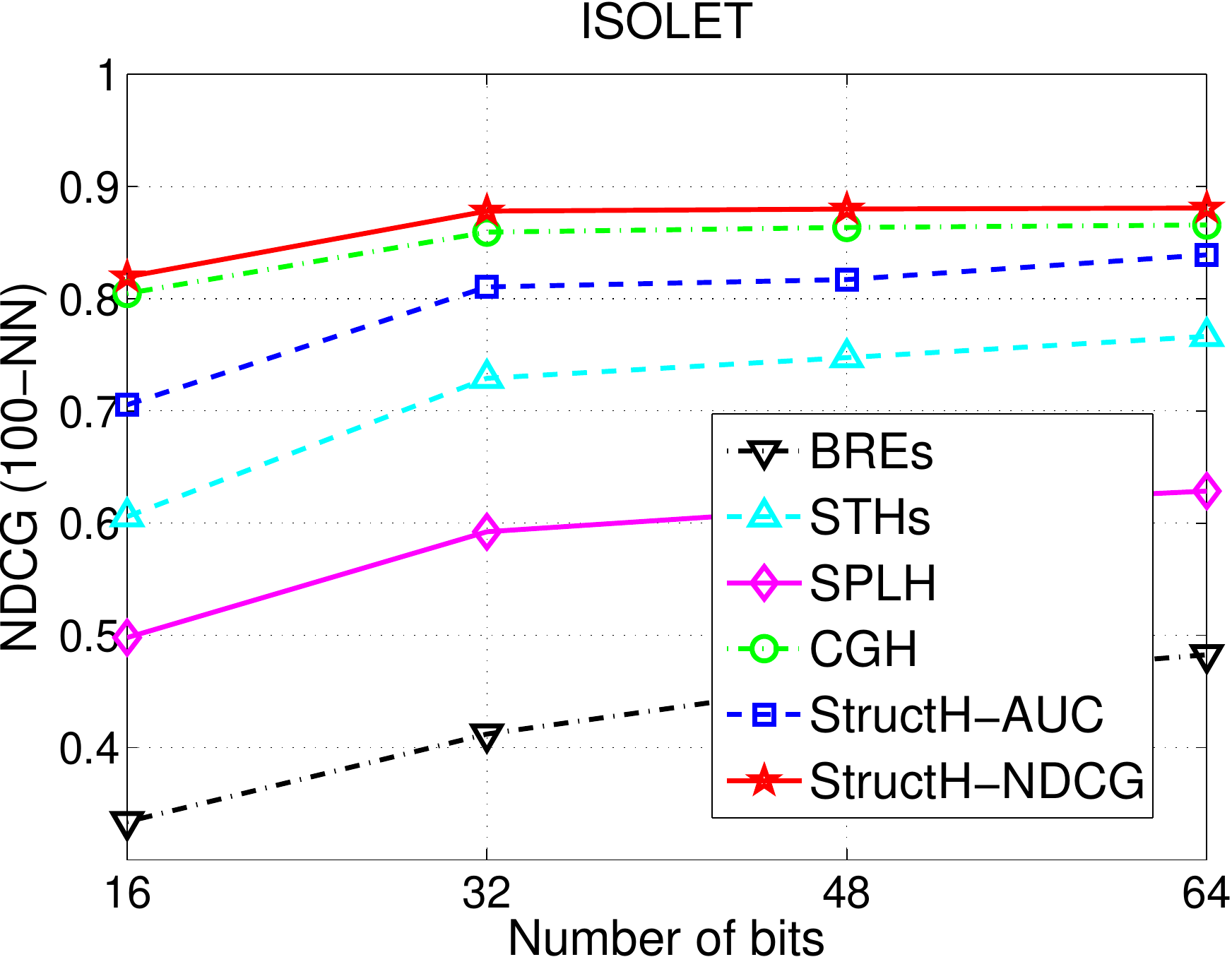}

 \caption{NDCG results on 3 datasets. Our StructHash performs the best.}
    \label{fig:ndcg_curve}
\end{figure*}

\begin{figure}[t]
    \centering

        \figcenter{{\includegraphics[height=0.25in]{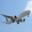}}}
                \colseperator
\figcenter{{\includegraphics[height=0.44in]{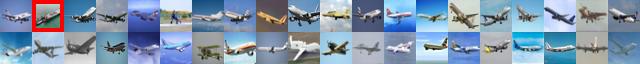}}}

        \figcenter{\includegraphics[height=0.25in]{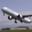}}
        \colseperator
        \figcenter{\includegraphics[height=0.44in]{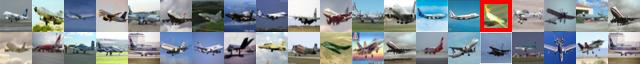}}

  \figcenter{\includegraphics[height=0.25in]{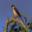}}
       \colseperator
       \figcenter{\includegraphics[height=0.44in]{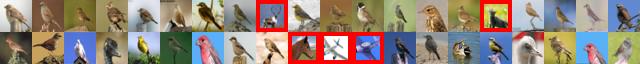}}

         \figcenter{\includegraphics[height=0.25in]{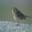}}
       \colseperator
       \figcenter{\includegraphics[height=0.44in]{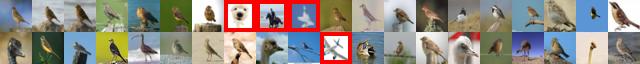}}

       \figcenter{\includegraphics[height=0.25in]{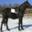}}
       \colseperator
       \figcenter{\includegraphics[height=0.44in]{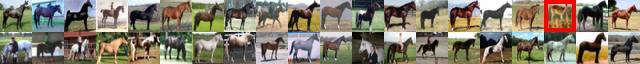}}

       \figcenter{\includegraphics[height=0.25in]{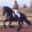}}
       \colseperator
       \figcenter{\includegraphics[height=0.44in]{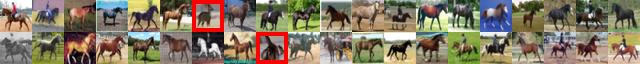}}

       \figcenter{\includegraphics[height=0.25in]{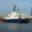}}
       \colseperator
       \figcenter{\includegraphics[height=0.44in]{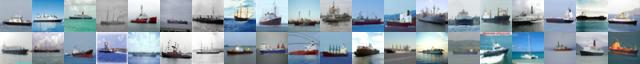}}

       \figcenter{\includegraphics[height=0.25in]{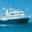}}
       \colseperator
       \figcenter{\includegraphics[height=0.44in]{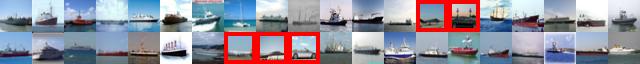}}

    \caption{Some ranking examples of our method. The first column shows query images, and the rest are retrieved images. False predictions are marked by red boxes.}
    \label{fig:examples_cifar}
\end{figure}

\begin{table*}[t]
\caption{Results using ranking measures of Precision-at-K, Mean Average Precision and Precision-Recall (64 bits).
 We compare our method using AUC (StructH-A) and NDCG (StructH-N) loss functions with other supervised and un-supervised methods. Our method using NDCG loss performs the best on these measures}
\centering
\resizebox{.9\linewidth}{!}
  {
  \begin{tabular}{ c || c c | c c c c | c c c c c}
\hline
Dataset & StructH-N	& StructH-A	&CGH	&SPLH	&STHs	&BREs	&ITQ	&SPHER	&MDSH	&AGH	&LSH \\
\hline
& \multicolumn{11}{c}{Precision-at-K ($K=100$)} \\ \hline
STL10	&\bf 0.431	&0.376	&0.376	&0.396	&0.208	&0.279	&0.325	&0.303	&0.298	&0.301	&0.222\\
USPS	&\bf 0.903	&0.894	&0.898	&0.805	&0.667	&0.755	&0.780	&0.730	&0.698	&0.711	&0.637\\
MNIST	&0.849	&0.807	&\bf 0.862	&0.797	&0.579	&0.790	&0.842	&0.788	&0.100	&0.780	&0.540\\
CIFAR	&0.336	&0.259	&0.261	&\bf 0.354	&0.174	&0.264	&0.301	&0.286	&0.270	&0.281	&0.164\\
ISOLET	&\bf 0.875	&0.844	&0.859	&0.604	&0.755	&0.448	&0.589	&0.477	&0.493	&0.493	&0.370\\
\hline
& \multicolumn{11}{c}{Mean Average Precision (mAP)} \\ \hline
STL10	&\bf 0.331	&0.326	&0.322	&0.299	&0.155	&0.211	&0.233	&0.193	&0.178	&0.162	&0.162\\
USPS	&\bf 0.868	&0.851	&0.848	&0.689	&0.456	&0.582	&0.566	&0.451	&0.405	&0.333	&0.418\\
MNIST	&\bf 0.802	&0.790	&0.789	&0.684	&0.397	&0.558	&0.585	&0.510	&0.119	&0.505	&0.343\\
CIFAR	&0.294	&\bf 0.300	&0.298	&0.289	&0.147	&0.204	&0.215	&0.204	&0.181	&0.201	&0.149\\
ISOLET	&\bf 0.836	&0.796	&0.815	&0.518	&0.653	&0.340	&0.484	&0.357	&0.348	&0.298	&0.267\\
\hline
& \multicolumn{11}{c}{Precision-Recall} \\ \hline
STL10	&\bf 0.267	&0.248	&0.248	&0.246	&0.130	&0.181	&0.200	&0.174	&0.164	&0.145	&0.138\\
USPS	&\bf 0.776	&0.760	&0.760	&0.609	&0.401	&0.520	&0.508	&0.424	&0.379	&0.326	&0.375\\
MNIST	&\bf 0.591	&0.574	&0.582	&0.445	&0.165	&0.313	&0.323	&0.246	&0.018	&0.197	&0.143\\
CIFAR	&0.105	&0.093	&0.091	&\bf 0.110	&0.042	&0.066	&0.074	&0.069	&0.064	&0.061	&0.042\\
ISOLET	&\bf 0.759	&0.709	&0.737	&0.445	&0.563	&0.301	&0.429	&0.321	&0.320	&0.275	&0.238\\
\hline
  \end{tabular}
  }
\label{tab:main-other}
\end{table*}

\begin{figure*}[t!]
    \centering

  \includegraphics[width=.32\linewidth]{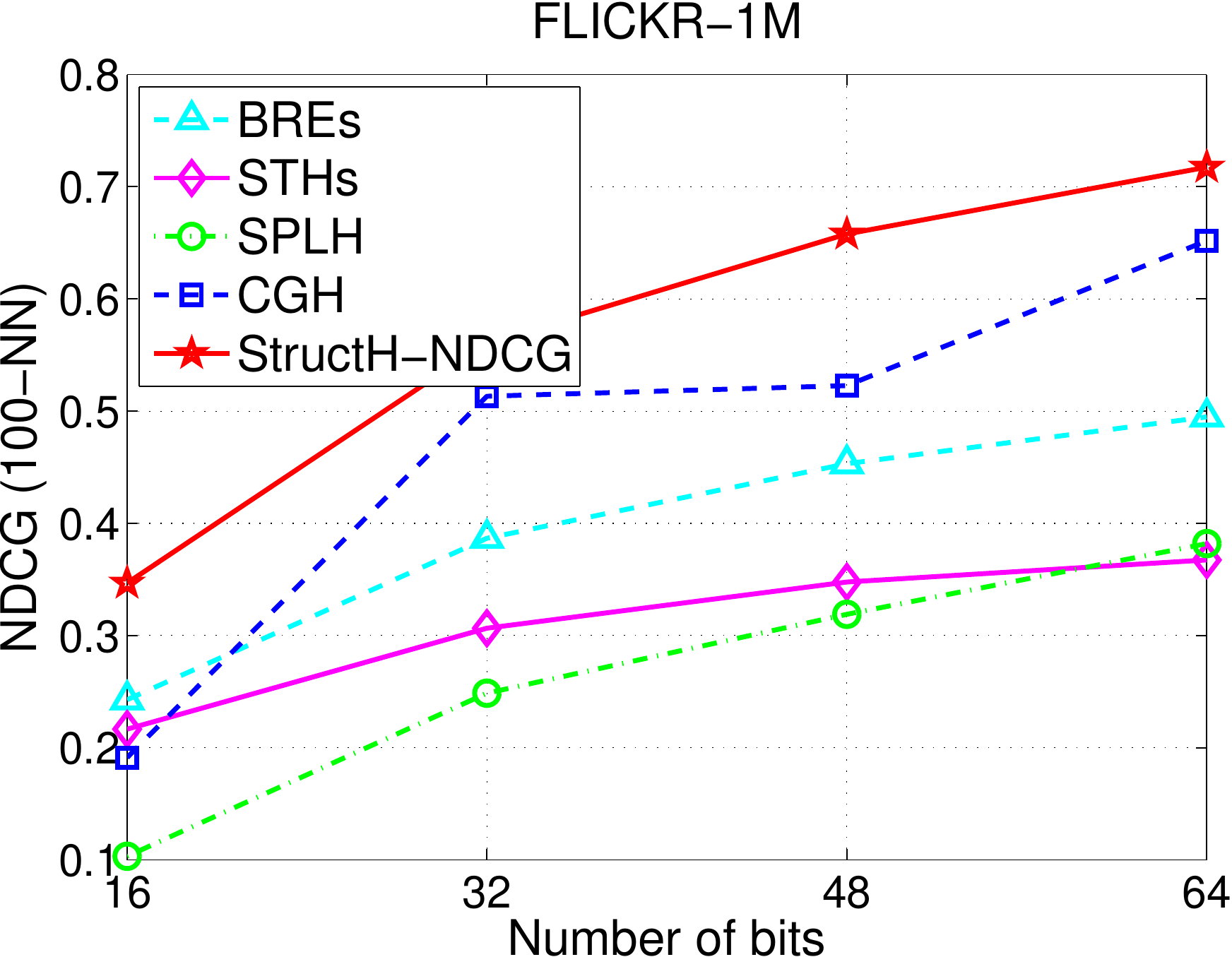}
  \includegraphics[width=.32\linewidth]{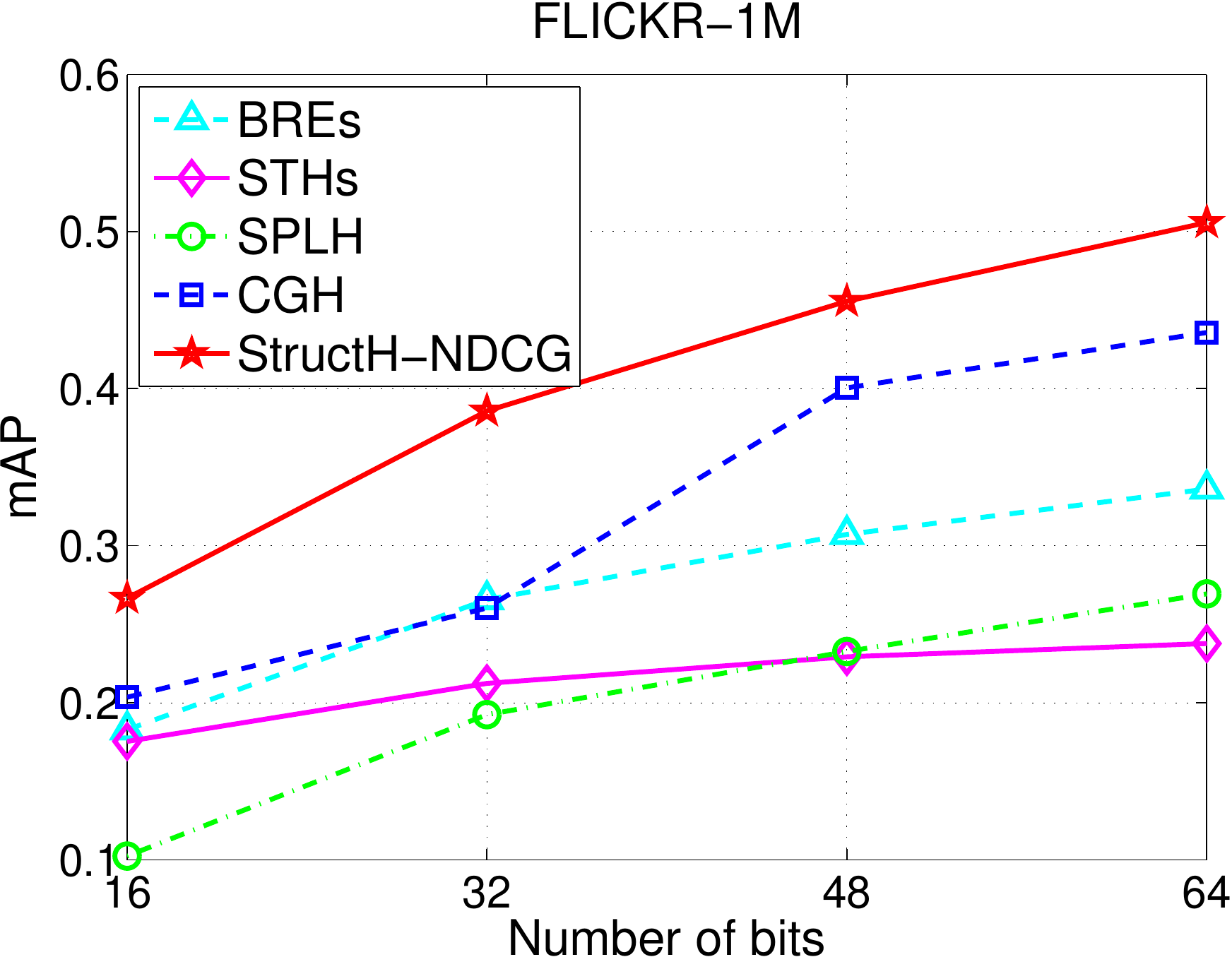}
  \includegraphics[width=.32\linewidth]{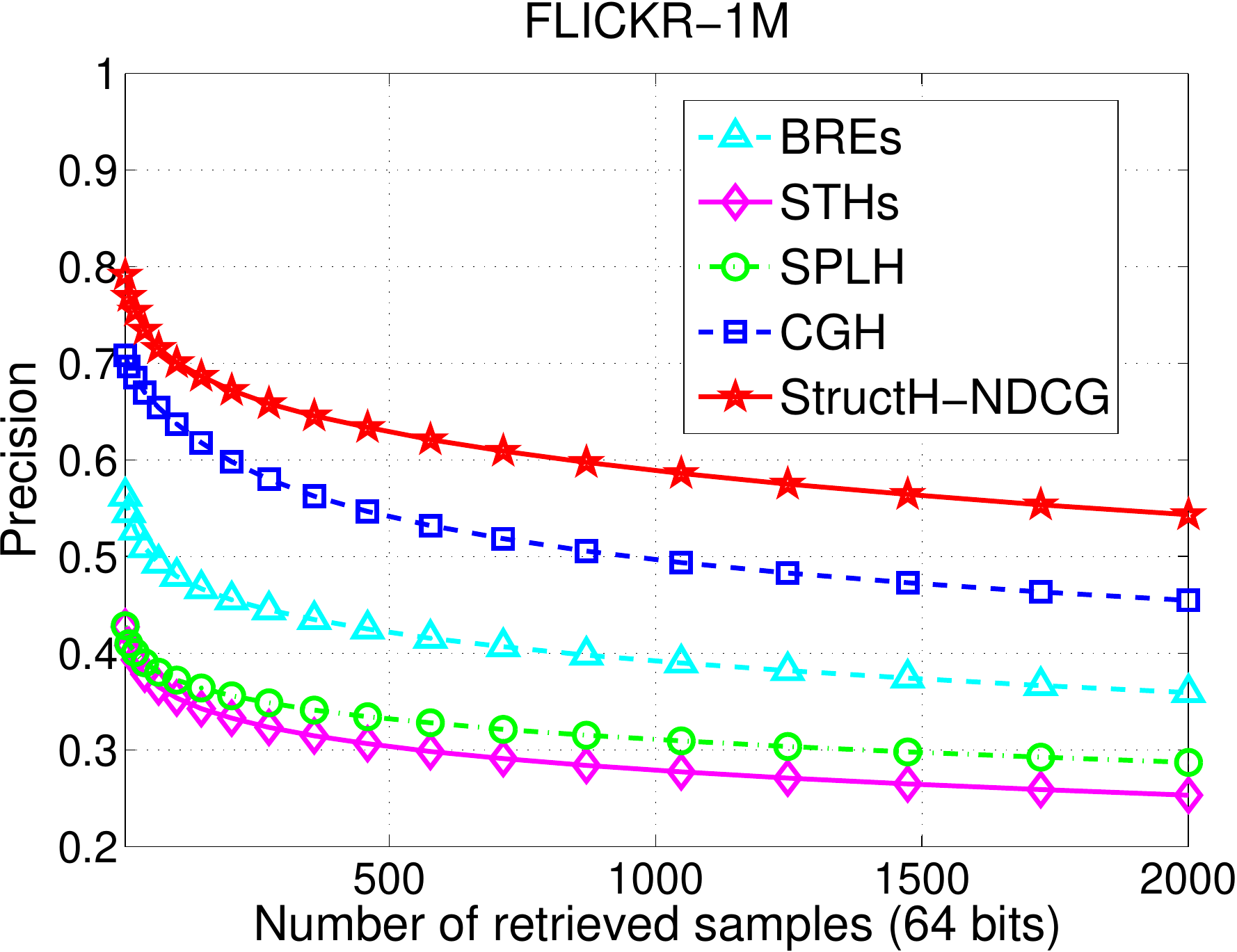}

  \includegraphics[width=.32\linewidth]{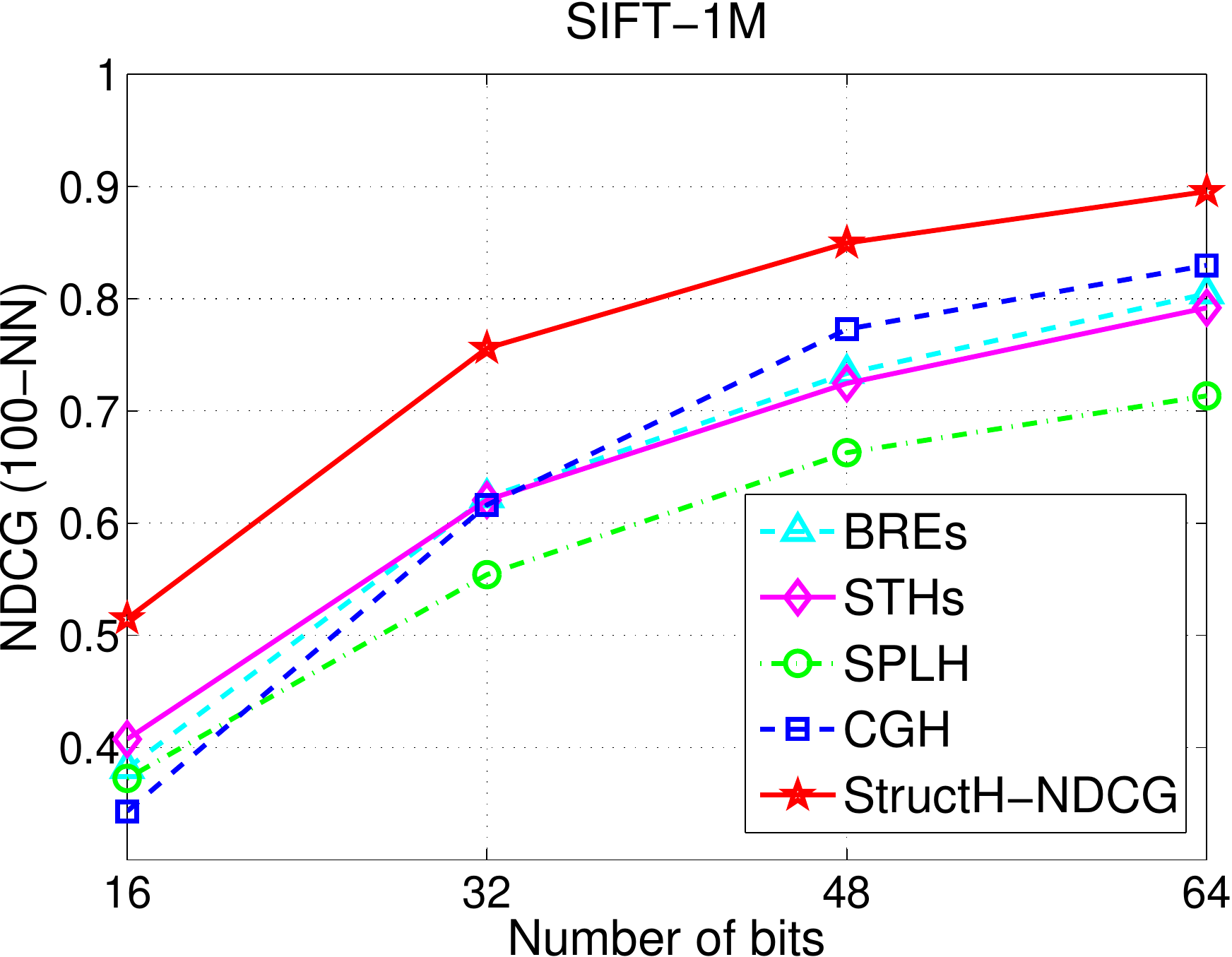}
  \includegraphics[width=.32\linewidth]{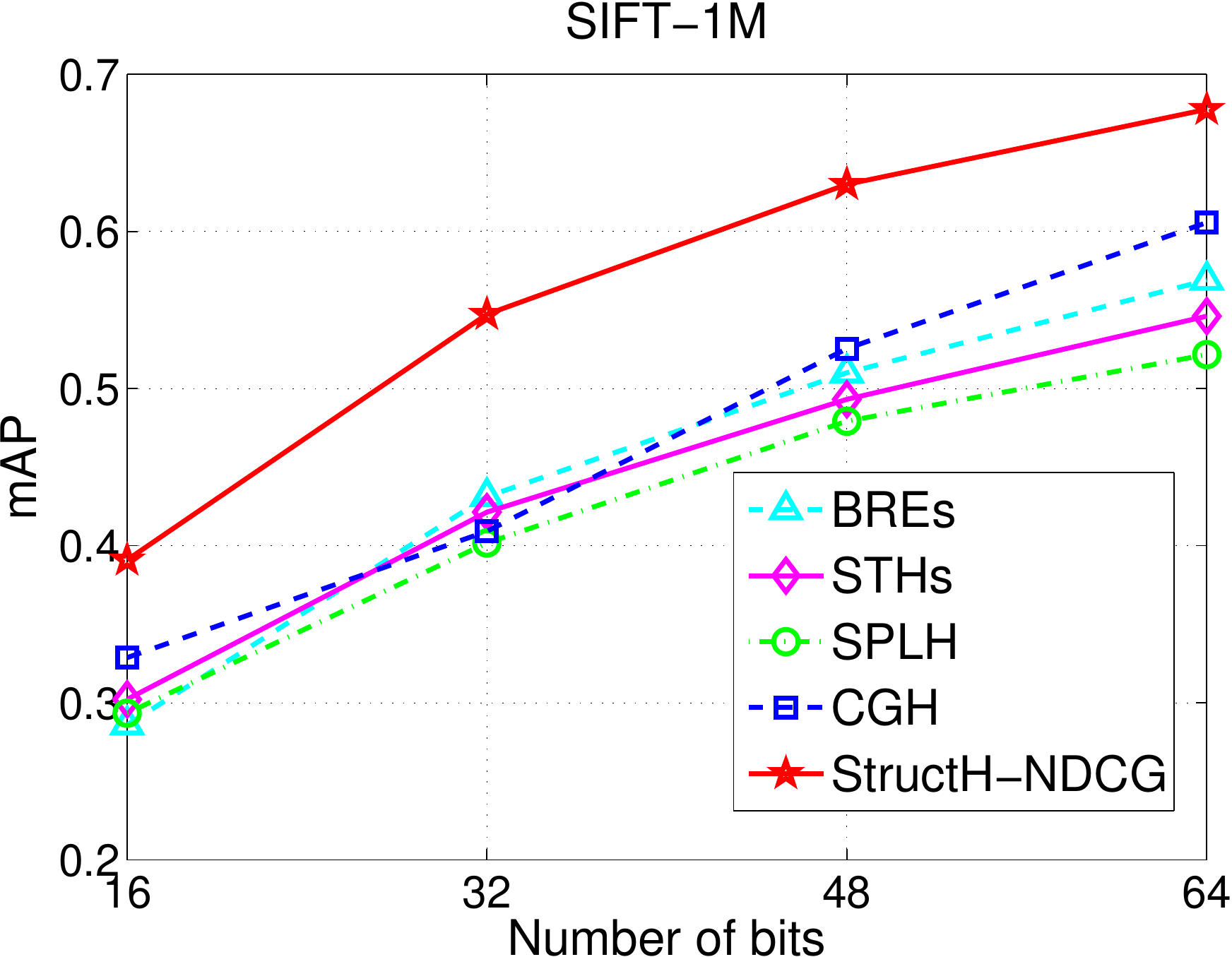}
  \includegraphics[width=.32\linewidth]{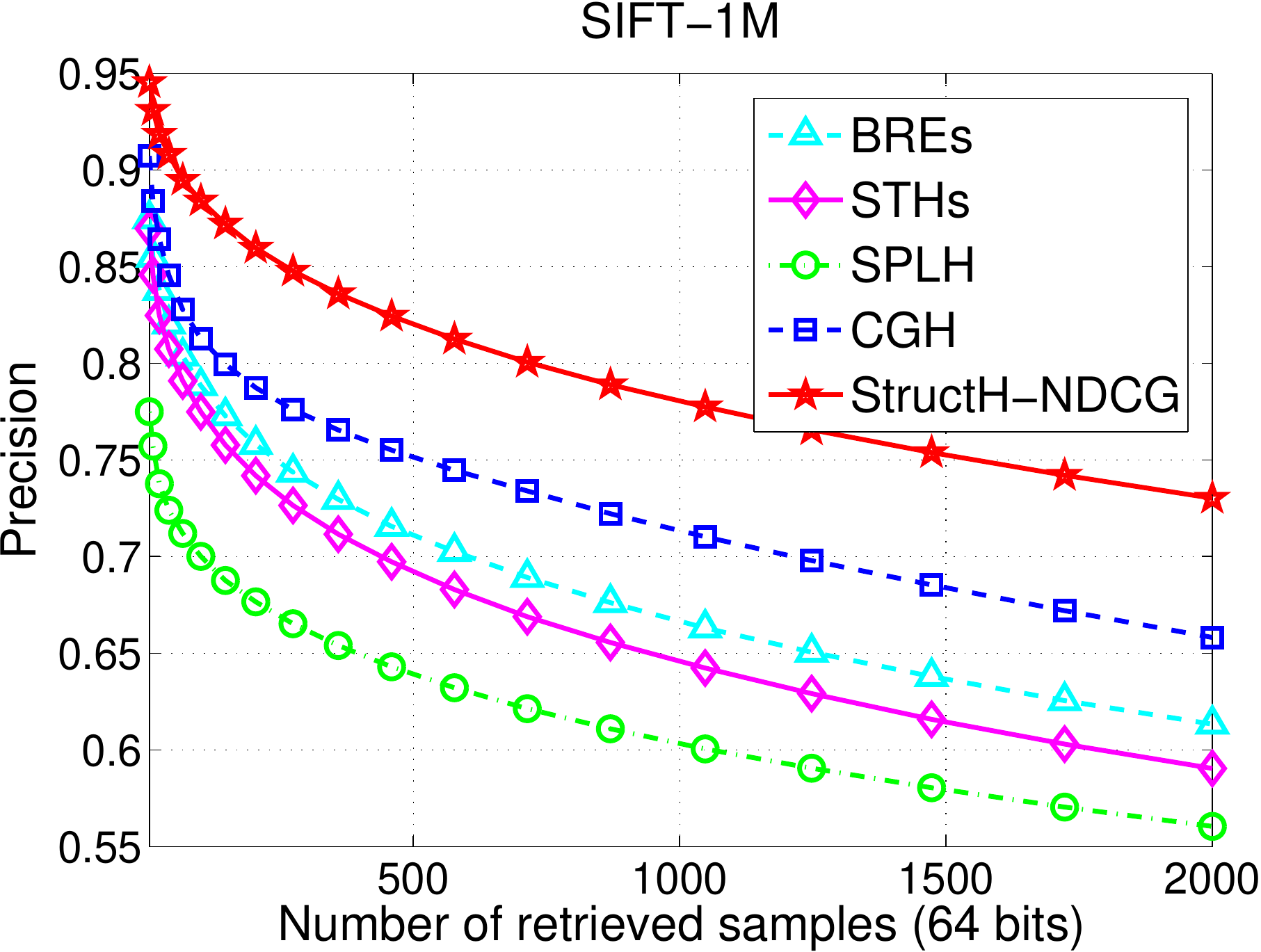}

  \includegraphics[width=.32\linewidth]{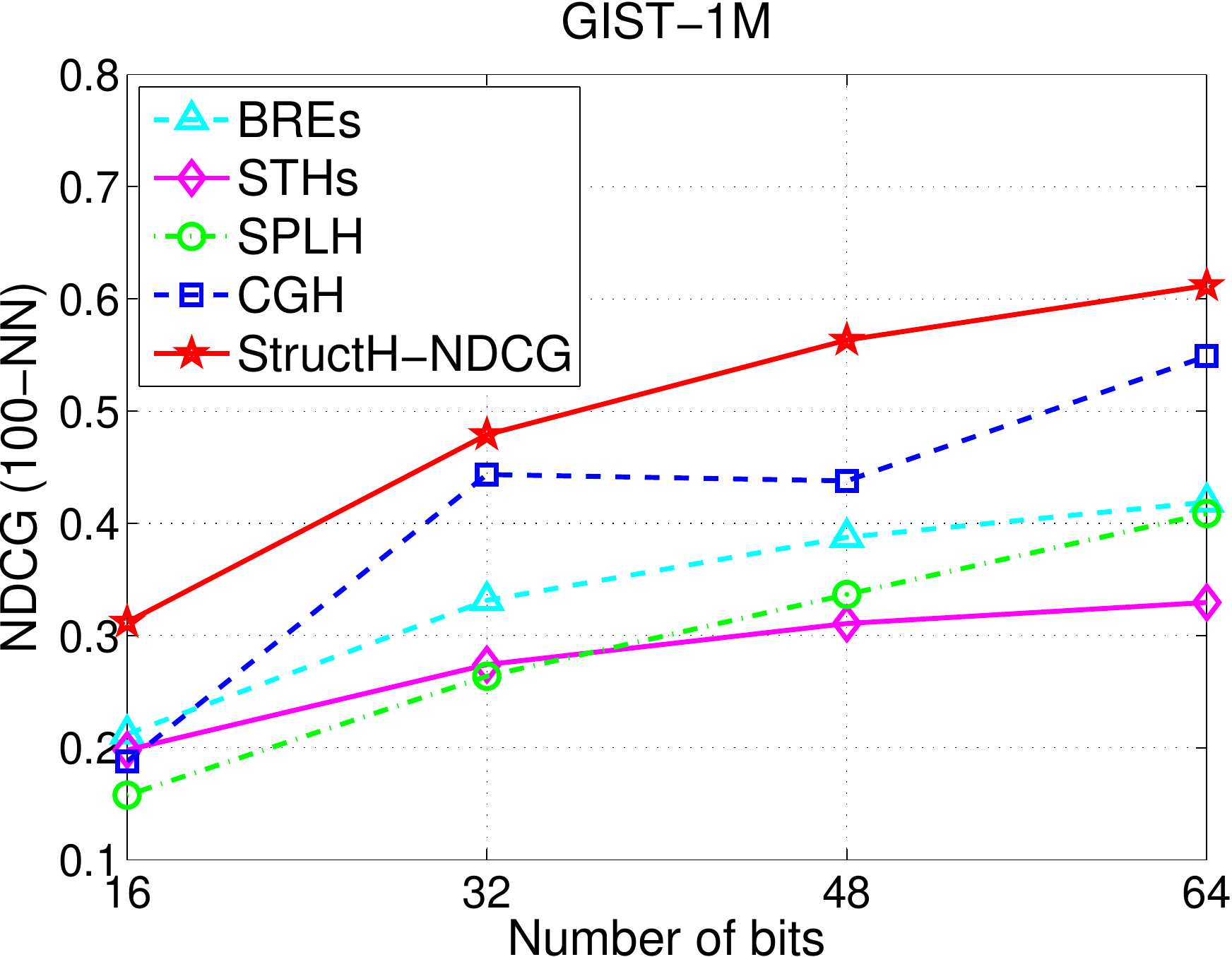}
  \includegraphics[width=.32\linewidth]{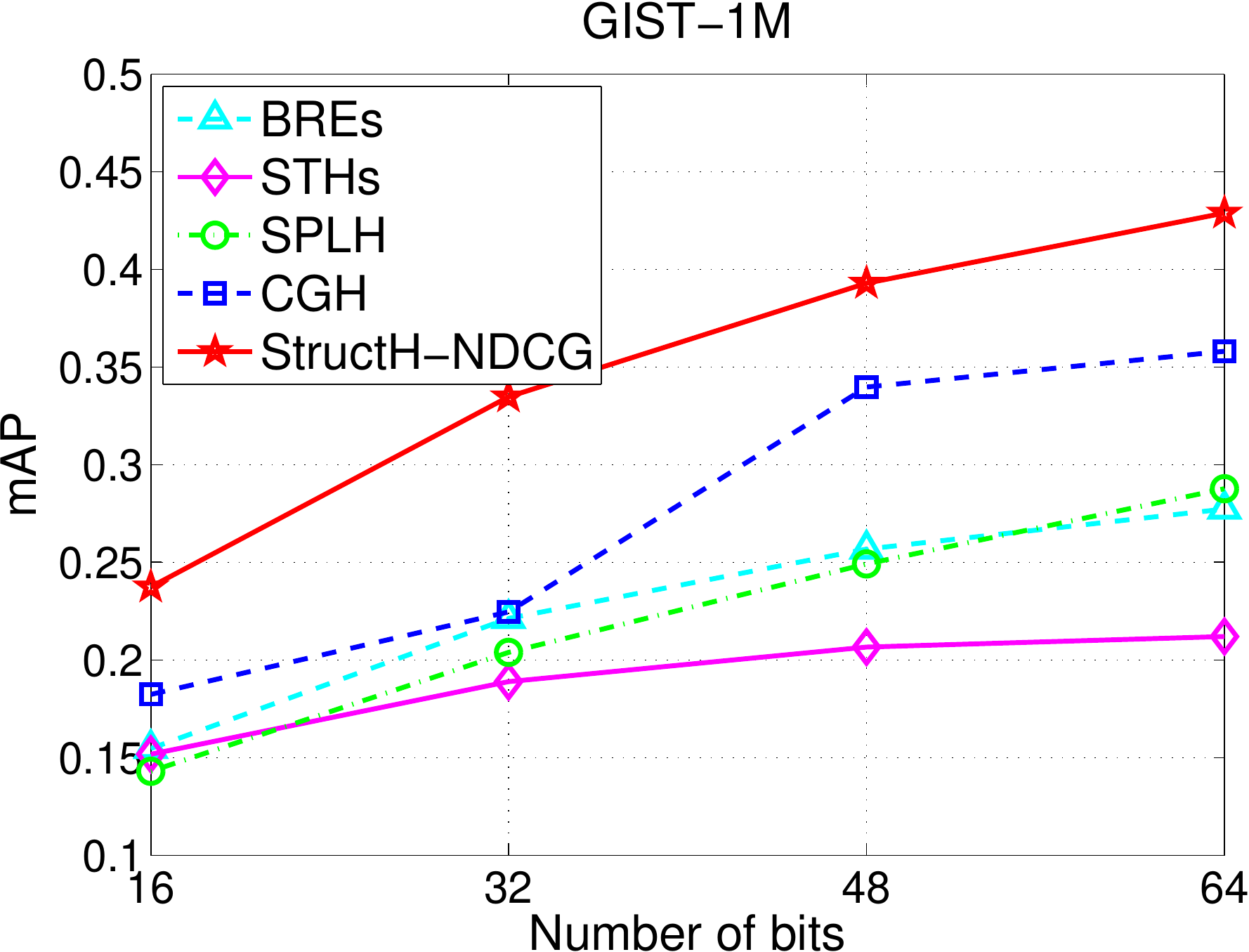}
  \includegraphics[width=.32\linewidth]{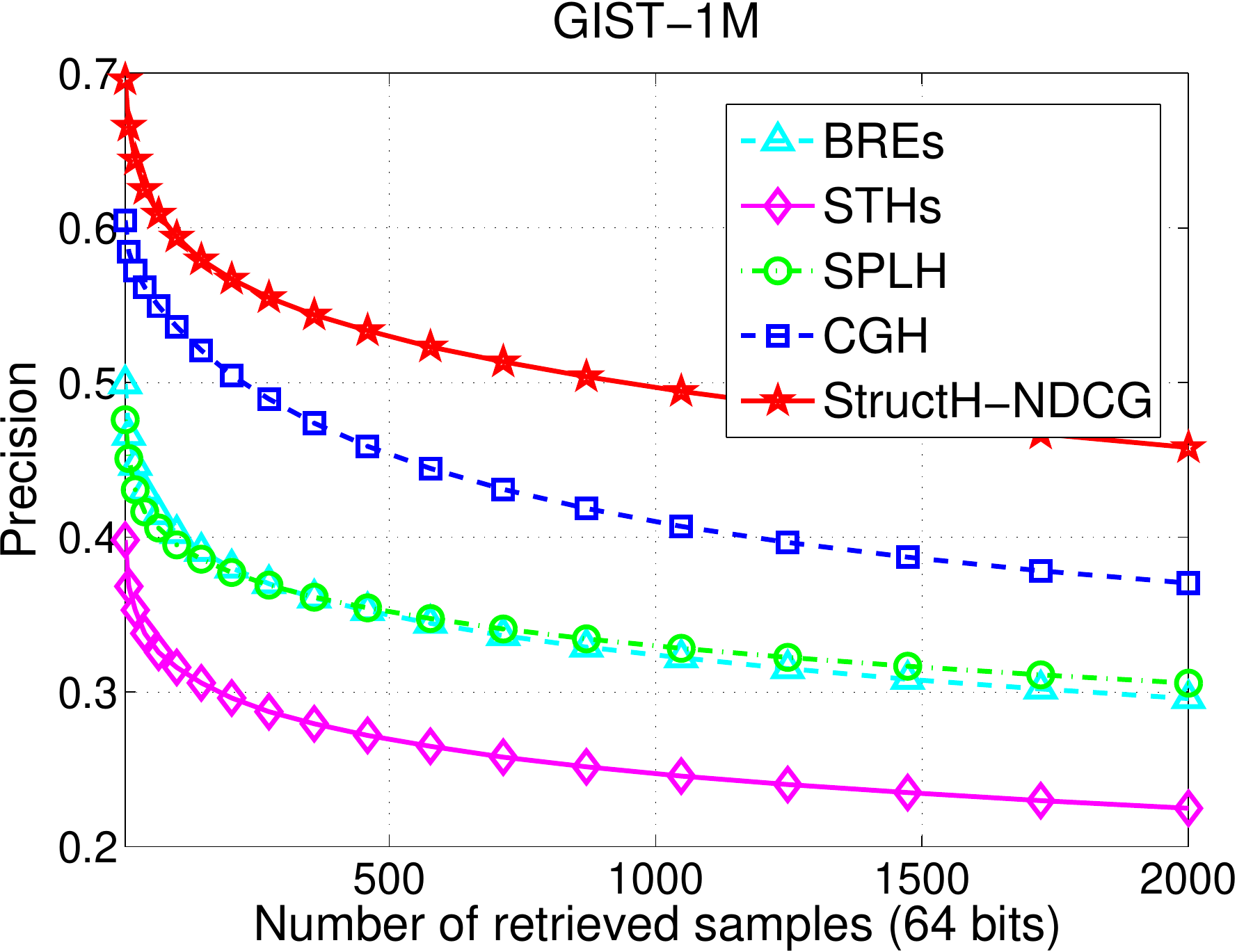}

  \includegraphics[width=.32\linewidth]{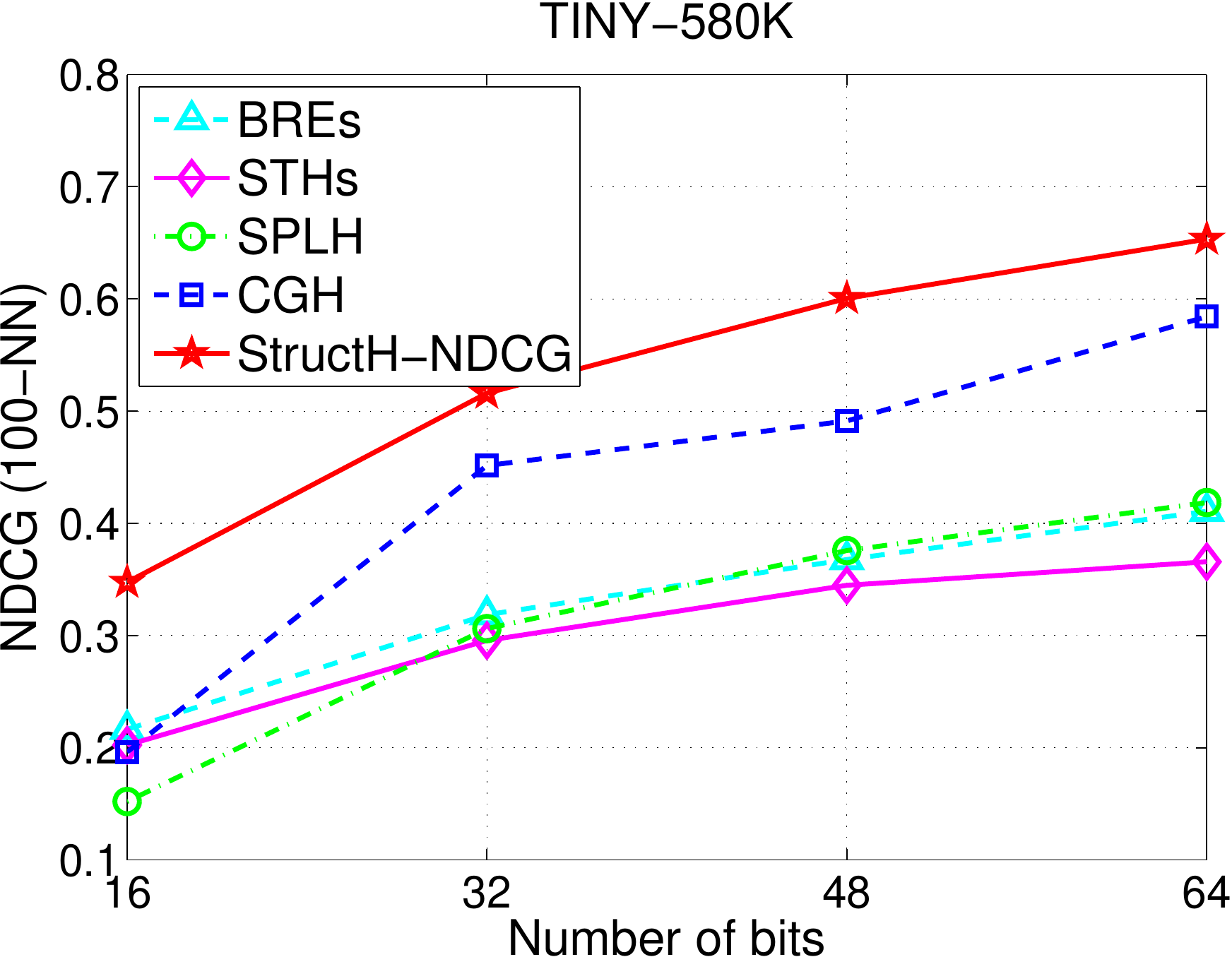}
  \includegraphics[width=.32\linewidth]{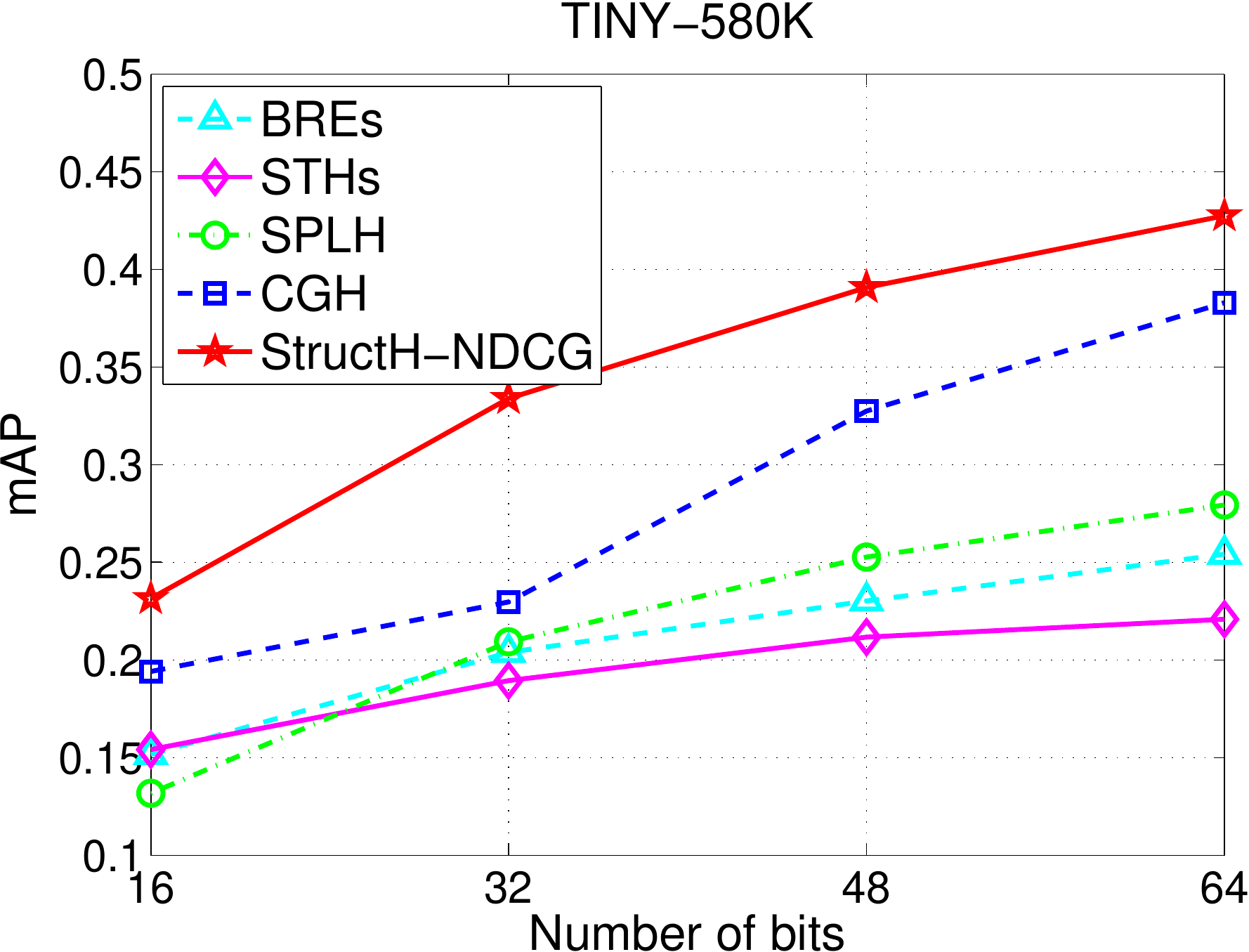}
  \includegraphics[width=.32\linewidth]{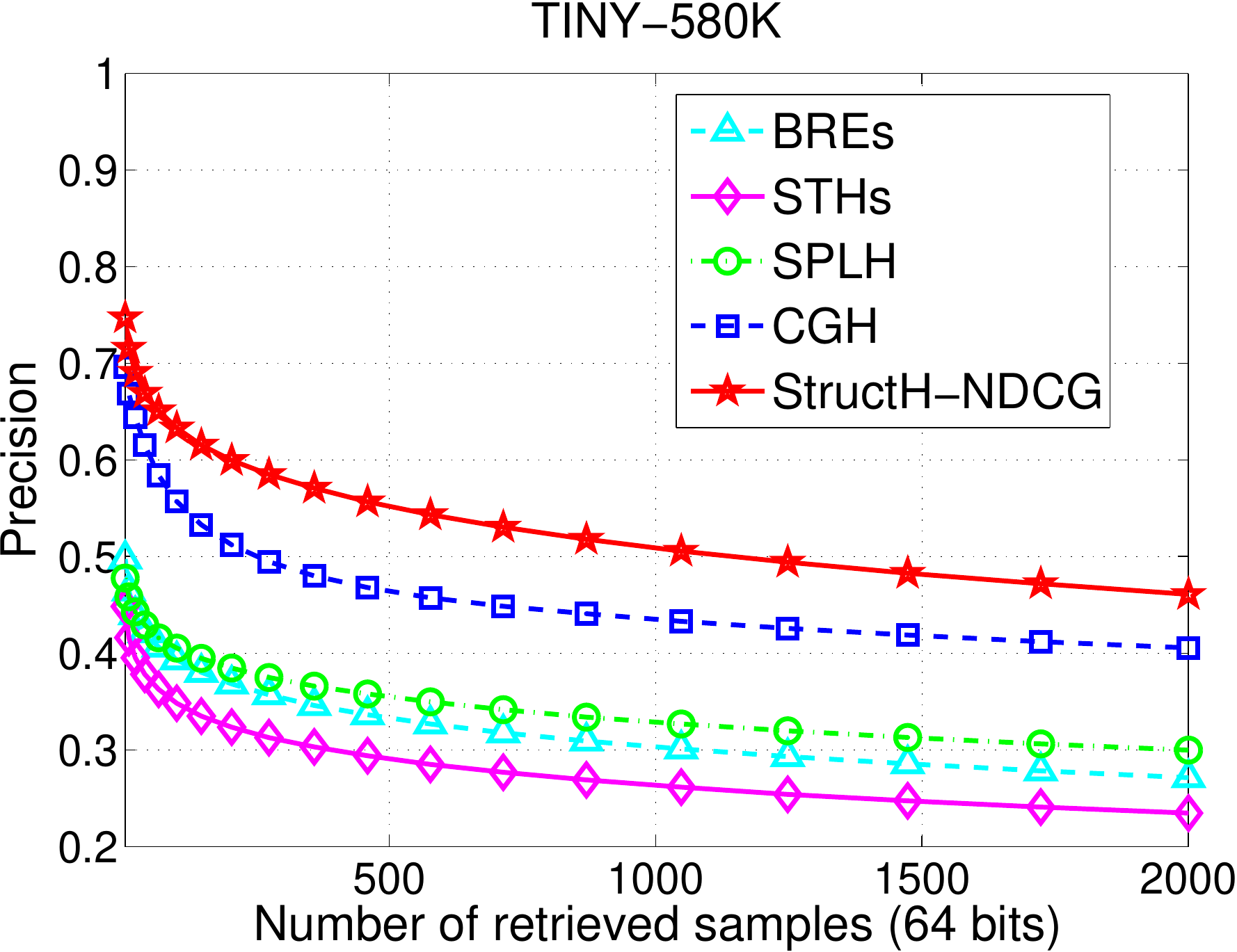}

 \caption{Results on 4 large datasests: Flickr-1M (1 million Flickr images), Sift-1M (1 million SIFT features), Gist-1M (1 million GIST features) and Tiny580K ($580,000$ Tiny image dataset). We compare with several supervised methods.
 The results of 3 measures (NDCG, mAP and precision of top-K neighbours) are show here.
 Our StructHash outperforms others in most cases.} \label{fig:large}
\end{figure*}

\begin{figure*}[t]
    \centering

  \includegraphics[width=.3\linewidth]{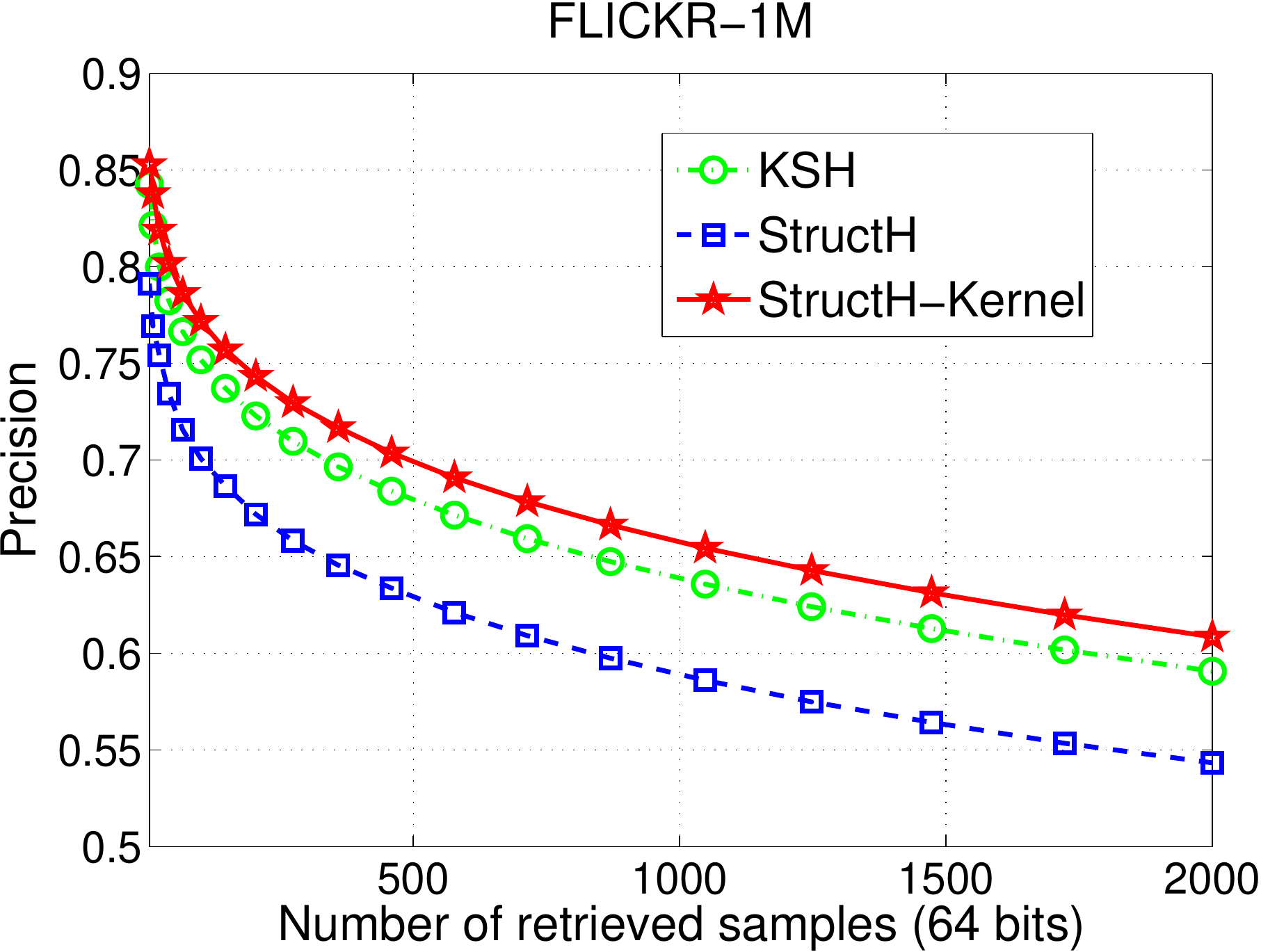}
  \includegraphics[width=.3\linewidth]{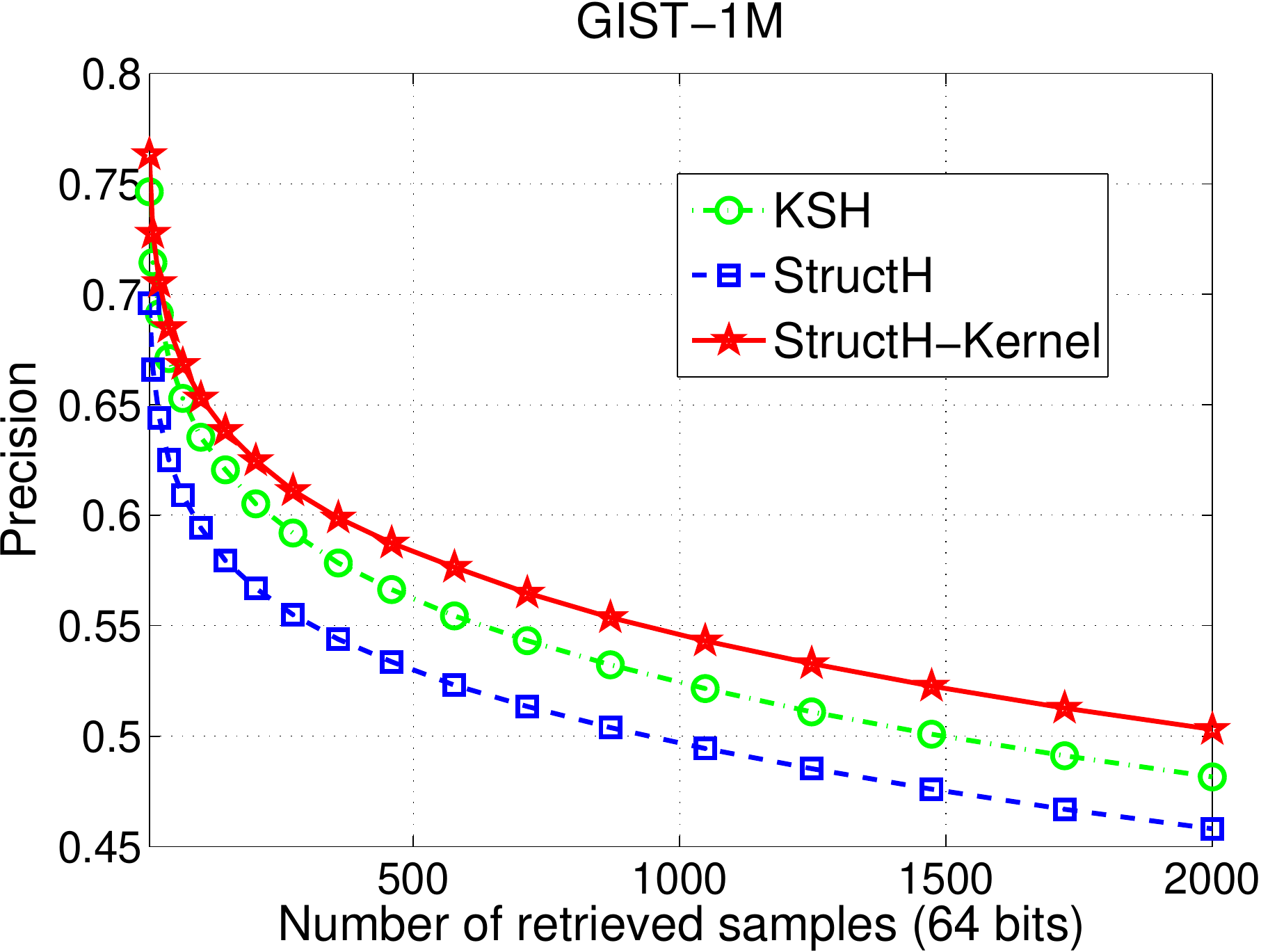}
  \includegraphics[width=.3\linewidth]{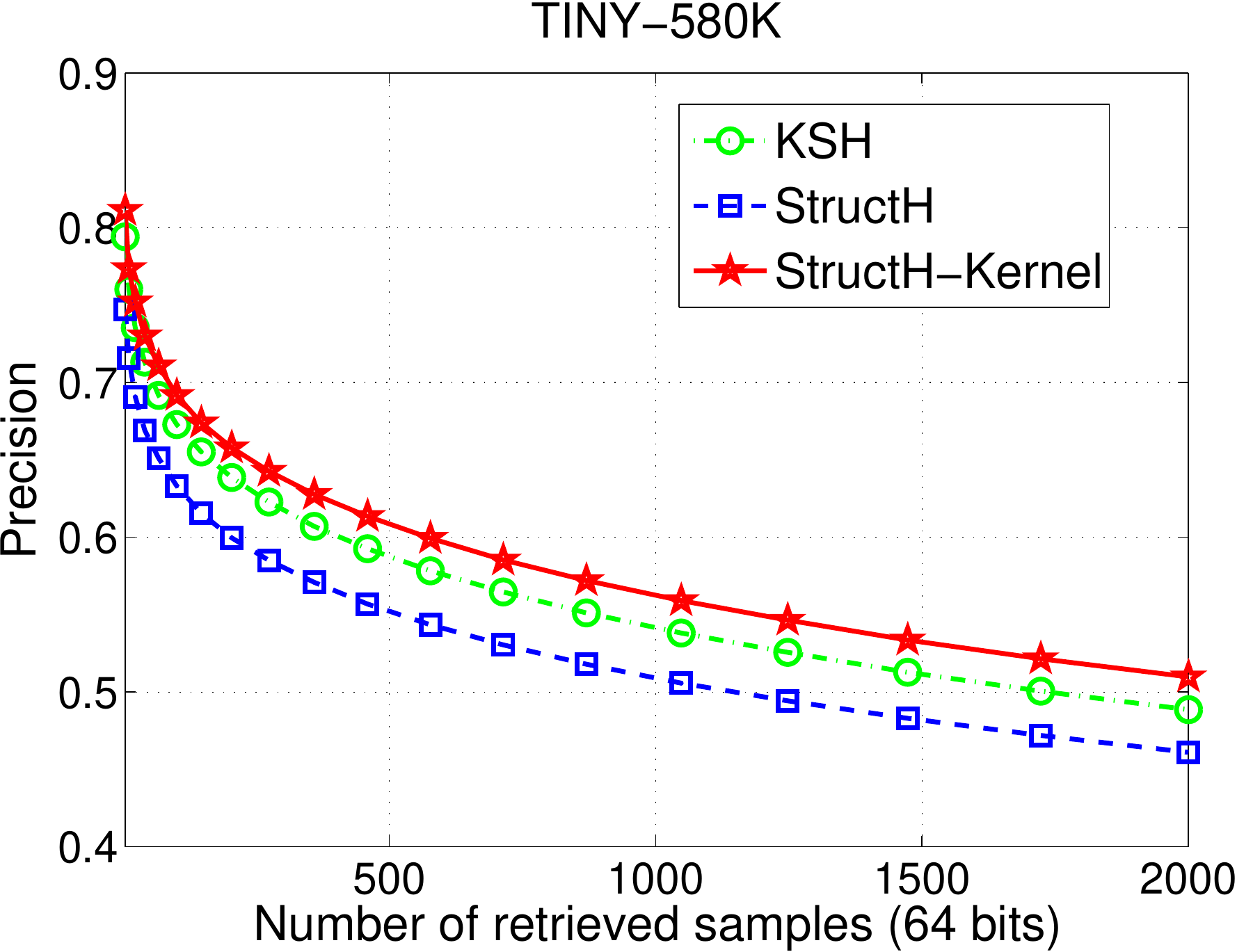}

 \caption{Comparison on large datasets of our kernel StructHash (StructHash-Kernel) with our non-kernel StructHash and the relevant method KSH \cite{KSH}. Our kernel version is able to achieve better results.} \label{fig:ksh}
\end{figure*}

\section{Experiments}

Our method is in the category of supervised method for learning compact binary codes. Thus
we mainly compare with 4 supervised methods: column
generation hashing (CGH) \cite{ICML13a},supervised binary reconstructive embeddings
(BREs) \cite{kulis2009learning}, supervised self-taught hashing (STHs)
\cite{zhangSTHs}, semi-supervised sequential projection learning
hashing (SPLH) \cite{wang2010semi}.

For comparison, we also run some unsupervised methods:
locality-sensitive hashing (LSH) \cite{Gionis1999}, anchor graph hashing
(AGH) \cite{liu2011hashingGraphs}, spherical hashing (SPHER) \cite{jae2012},
multi-dimension spectral hashing (MDSH) \cite{MDSH}, and iterative quantization (ITQ) \cite{gong2012iterative}.
We carefully follow the original authors' instruction for parameter setting.
For SPLH, the regularization parameter is
picked from $0.01$ to $1$. We use the hierarchical variant of AGH. The
bandwidth parameters of Gaussian affinity in MDSH is set as $\sigma=t\bar{d}$. Here $\bar{d}$ is the average
Euclidean distance of top $100$ nearest neighbours and $t$ is picked
from $0.01$ to $50$.
For supervised training of our \sh and CGH, we use 50
relevant and 50 irrelevant examples
to construct similarity information for each data point.

We use 9 datasets for evaluation, including  one UCI dataset: ISOLET, 4 image datasets: CIFAR10
\footnote{http://www.cs.toronto.edu/\~{}kriz/cifar.html},
STL10\footnote{http://www.stanford.edu/\~{}acoates/stl10/},
MNIST, USPS, and another 4 large image datasets:
Tiny-580K\cite{gong2012iterative}, Flickr-1M\footnote{http://press.liacs.nl/mirflickr/}, SIFT-1M \cite{wang2010semi} and GIST-1M\footnote{http://corpus-texmex.irisa.fr/}.
CIFAR10 is a subset of the 80-million tiny images and STL10 is a subset of Image-Net.
Tiny-580K consists of $580,000$ tiny images.
Flick-1M dataset consists of 1 million thumbnail images.
SIFT-1M and GIST-1M datasets contain 1 million SIFT and GIST features respectively.

We follow a common
setting in many supervised methods \cite{kulis2009learning,KSH,ICML13a} for hashing evaluation.
For multi-class datasets, we use class labels to
define the relevant and irrelevant semantic neighbours by label
agreement.
For large datasets: Flickr-1M, SIFT-1M, GIST-1M and Tiny-580K, the semantic ground truth is defined according to the
$\ell_2$ distance \cite{wang2010semi}.
Specifically, a data point is labeled as a
relevant data point of the query if it lies in the top 2 percentile
points in the whole dataset.
We generated GIST features for all image datasets except MNIST and USPS.
we randomly select 2000 examples for testing
queries, and the rest is used as database.
We sample 2000 examples from the database as training data for learning models.
For large datasets, we use 5000 examples for training.
To evaluate the performance of compact bits, the maximum bit length is set to 64,
as similar to the evaluation settings
in other supervised hashing methods \cite{kulis2009learning,ICML13a}.

We report the result of the NDCG measure in Table \ref{tab:main}.
We compare our StructHash using AUC and NDCG loss functions with other supervised and un-supervised methods.  Our method using NDCG loss function performs the best in most cases.
We also report the result of other common measures in Table \ref{tab:main-other},
including the result of Precision-at-K, Mean Average Precision (mAP) and Precision-Recall.
Precision-at-K is the proportion of true relevant data points in the returned top-K results.
The Precision-Recall curve measures the overall performance in all positions of the prediction ranking, which
is computed by varying the number of nearest neighbours.
It shows that our method generally performs better than other methods on these evaluation measures.
As described before, compared to the AUC measure which is position insensitive,
the NDCG measure assigns different importance on ranking positions, which is closely related to many other position sensitive ranking measures (e.g., mAP).
As expected, the result shows that on the Precision-at-K, mAP and Precision-recall measures,
optimizing the position sensitive NDCG loss performs better than the AUC loss.
The triplet loss based method CGH actually helps to reduce the AUC loss.
This may explain
the reason that our method, which aims to reduce the NDCG loss, is able to outperform CGH in
these measures.
We also plot the NDCG results on several datasets in Fig. \ref{fig:ndcg_curve} by varying the number of bits.
Some retrieval examples are shown in Fig. \ref{fig:examples_cifar}.

We further evaluate our method on 4 large-scale datasets (Flickr-1M, SIFT-1M, GIST-1M and Tiny-580K).
The results of NDCG, mAP and the precision of top-K neighbours are shown in Fig. \ref{fig:large}.
The NDCG and mAP results are shown by varying the number of bits.
The precision of top-K neighbours is shown by varying the number of retrieved examples.
In most cases, our method outperforms other competitors.
Our method with NDCG loss function succeeds to achieve good performance both on NDCG and other measures.

Applying the kernel technique in KLSH \cite{KLSH} and KSH \cite{KSH} further improves the performance of our method. As describe in \cite{KSH}, we perform a pre-processing step to generate the kernel mapping features: we randomly select a number of support vectors (300) then compute the kernel response on data points as input features.
Note that here we simply follow KSH for the kernel parameter setting.
We evaluate this kernel version of our method in Fig. \ref{fig:ksh} and compare to KSH.
Our kernel version is able to achieve better results.

\section{Conclusion and future work}
We have developed a hashing framework that allows us to directly optimize multivariate performance measures.
The fact that the proposed method outperforms comparable
hashing approaches is to be expected, as it more directly optimizes
the required loss function.
It is anticipated that the success of the approach may lead to a range of new
hashing-based applications with task-specified targets. Extension to make use of the power
of more sophisticated hash functions such as kernel functions or decision trees
is of future work.

\bibliographystyle{splncs03}
\bibliography{structhash}

\end{document}